\documentclass[journal,twoside]{IEEEtran}
\usepackage[ruled,vlined]{algorithm2e}
\usepackage{graphicx,amssymb,hyperref}
\usepackage{epstopdf}
\usepackage{booktabs}
\usepackage{color}
\usepackage{multirow}
\usepackage{amsmath}
\usepackage{subfigure}
\usepackage{bm}
\usepackage{url}
\usepackage{bbding}
\usepackage{indentfirst}
\usepackage{bbm}
\usepackage[T1]{fontenc}
\usepackage{cite}

\usepackage{fancyhdr}

\begin{document}

\title{Reasoning Graph Networks for Kinship Verification: from Star-shaped to Hierarchical}

\author{Wanhua~Li,~\IEEEmembership{Student~Member,~IEEE}, Jiwen~Lu,~\IEEEmembership{Senior~Member,~IEEE}, Abudukelimu~Wuerkaixi,
Jianjiang~Feng, \IEEEmembership{Member,~IEEE}, and Jie~Zhou,~\IEEEmembership{Senior~Member,~IEEE}
\thanks{
This work was supported in part by the National Key Research and Development Program of China under Grant 2017YFA0700802, in part by the National Natural Science Foundation of China under Grant 61822603, Grant U1813218, Grant U1713214, and Grant 61672306, in part by Beijing Academy of Artificial Intelligence (BAAI), and in part by a grant from the Institute for Guo Qiang, Tsinghua University. \emph{(Corresponding author: Jie Zhou)}.

The authors are with the Beijing National Research Center for Information Science and Technology (BNRist), and the Department of Automation, Tsinghua University, Beijing, 100084, China.
E-mail: li-wh17@mails.tsinghua.edu.cn;~lujiwen@tsinghua.edu.cn;~wekxabdk17@mails.tsinghua.edu.cn;~jfeng@tsinghua.edu.cn;~jzhou@tsinghua.edu.cn.}

\markboth{IEEE Transactions on Image Processing}%
{Li~\MakeLowercase{\textit{et al.}}: Learning Hierarchical Reasoning Graph Networks for Kinship Verification}
}
\maketitle

\thispagestyle{fancy}
\fancyhead{}
\lhead{}
\lfoot{1941-0042~\copyright~2021 IEEE. Personal use is permitted, but republication/redistribution requires IEEE permission. See https://www.ieee.org/publications/rights/index.html for more information.}
\cfoot{}
\rfoot{}

\begin{abstract}
In this paper, we investigate the problem of facial kinship verification by learning hierarchical reasoning graph networks. Conventional methods usually focus on learning discriminative features for each facial image of a paired sample and neglect how to fuse the obtained two facial image features and reason about the relations between them. To address this, we propose a Star-shaped Reasoning Graph Network (S-RGN). Our S-RGN first constructs a star-shaped graph where each surrounding node encodes the information of comparisons in a feature dimension and the central node is employed as the bridge for the interaction of surrounding nodes. Then we perform relational reasoning on this star graph with iterative message passing. The proposed S-RGN uses only one central node to analyze and process information from all surrounding nodes, which limits its reasoning capacity. We further develop a Hierarchical Reasoning Graph Network (H-RGN) to exploit more powerful and flexible capacity. More specifically, our H-RGN introduces a set of latent reasoning nodes and constructs a hierarchical graph with them. Then bottom-up comparative information abstraction and top-down comprehensive signal propagation are iteratively performed on the hierarchical graph to update the node features. Extensive experimental results on four widely used kinship databases show that the proposed methods achieve very competitive results.

\end{abstract}
\IEEEpeerreviewmaketitle

\section{Introduction}
\IEEEPARstart
The human face contains rich information, such as age, gender, ethnicity, identity, and so on. Many facial image analysis problems including facial age estimation~\cite{dibekliouglu2015combining,li2019bridgenet}, emotion recognition~\cite{li2018occlusion}, gender classification~\cite{moghaddam2000gender}, and face recognition~\cite{ge2018low,deng2019arcface,ding2015multi} have been extensively studied for decades. As an emerging and interesting face related task, kinship verification aims to determine whether or not a kin relation exists for a given pair of facial images. Kinship verification is inspired by the biology finding~\cite{dal2010lateralization} that human facial appearance encodes important kin related cues. Although it is a quite difficult problem, continuous efforts~\cite{lu2013neighborhood,yan2014discriminative,BMVC2015_148,kohli2016hierarchical} have been devoted due to broad applications such as missing children searching~\cite{lu2013neighborhood}, automatic album organization~\cite{zhou2012gabor}, children adoptions~\cite{BMVC2015_148}, and social media-based analysis~\cite{dehghan2014look}.

Generally, there are two main stages for kinship verification: face representation and face matching. Face representation aims to extract discriminative features for each facial image, and face matching is to design models to fuse two extracted features and predict the genetic relationship between them. Several challenges prevent it from being deployed in any real-world application. First, as other face-related tasks~\cite{ge2018low,li2018occlusion,li2019bridgenet,moghaddam2000gender},  kinship verification is also confronted with a large variation in facial appearance caused by pose, scale, expression, illumination, etc. The large variation makes learning discriminative features quite challenging. Second, facial kinship verification has to discover the genetic relations between two samples from facial appearance. The challenge is even bigger since kinship verification has to discover the hidden similarity inherited by genetic relations between different identities. Cross-identity relational reasoning naturally leads to a much larger gap in the facial appearance of intra-class samples, especially when there are significant age gaps and gender differences.

\begin{figure}[t]
  \centering
  \includegraphics[width=0.9\linewidth]{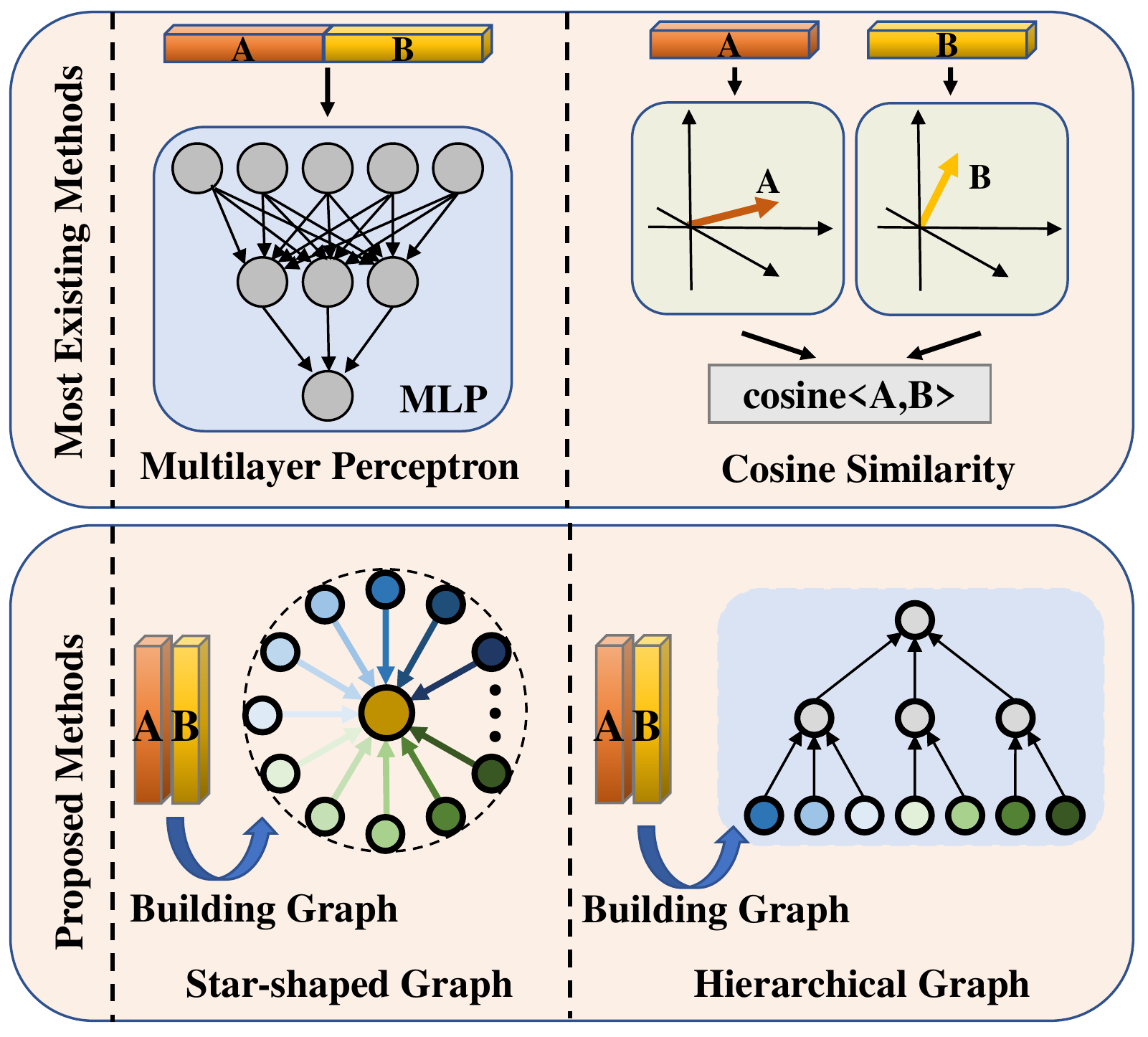}
  \caption{Main differences between our approaches and other methods. Most existing methods usually focus on the face representation stage and simply apply a similarity metric like cosine similarity, or a multilayer perceptron layer to the extracted facial image features in the face matching stage, which couldn't fully exploit the hidden genetic relations. By contrast, our methods concentrate on the face matching stage and perform relational reasoning on the constructed star-shaped or hierarchical reasoning graphs.}
  \label{fig:introduction}
  \vspace{-0.5cm}
\end{figure}

To address these challenges, many methods~\cite{lu2013neighborhood,BMVC2015_148} have been proposed over the past few years. Most of them focus on the face representation stage and aim to learn discriminative features for each image of a paired facial sample. For example, metric-learning based methods usually project initial features into a latent feature space by learning a metric to get discriminative projected features~\cite{fan2020efficientpr}. Lu \emph{et al.}~\cite{lu2013neighborhood} proposed the neighborhood repulsed metric learning method to pull intra-class samples with a kinship relation as close as possible and push interclass samples lying in a neighborhood as far as possible. Hu \emph{et al.}~\cite{hu2017local} presented a large-margin multi-metric learning method for kinship verification, which jointly learns multiple distance metrics to exploit the complementary information. Encouraged by the success of deep convolutional neural networks (CNN) in many vision tasks such as image recognition~\cite{he2016deep}, object detection~\cite{faster17pami,ouyang17deepid}, and face recognition~\cite{deng2019arcface,ding2017trunk}, some deep learning methods~\cite{BMVC2015_148,dibeklioglu2017visual} have been proposed for kinship verification in recent years. Zhang \emph{et al.}~\cite{BMVC2015_148} first proposed using a CNN to extract deep features for kinship verification to fully exploit the powerful feature representation ability of deep learning. However, few works investigate the face matching stage and consider how to fuse the two extracted features. In the face matching stage, most existing methods either send the concatenated features into a Multilayer Perceptron (MLP) layer or calculate the cosine similarity between the two features, as depicted in Fig. \ref{fig:introduction}. All these methods can not effectively model the relations between two extracted features to reason about the existence of kinship between them.

In this work, we focus on the face matching stage and consider how to compare and fuse two facial image features to infer the genetic relations between them. Although kinship verification is very challenging even for humans, sometimes people can make accurate predictions by first comparing some biological characteristics of two individuals, such as eye color, nose size, and cheekbone shape, and then making a comprehensive analysis based on these comparison results. Inspired by the above reasoning process of humans, we try to explicitly model it in the face matching period. Having obtained two extracted facial image features, we consider that different dimensions of features encode different kinship related information. We can make predictions by first comparing features dimensionally and then fusing these dimension by dimension comparison results.

To this end, we first present a Star-shaped Reasoning Graph Network (S-RGN), which first constructs a star graph for two extracted image features and then performs relational reasoning on this graph to effectively exploit the hidden kin relations between two individuals.
To fuse the comparison information in all feature dimensions, a central node is employed in our S-RGN, which may limit the reasoning ability and flexibility of our model. Therefore, we further propose a Hierarchical Reasoning Graph Network (H-RGN), where a set of latent nodes is introduced to construct a hierarchical reasoning graph. We adopt a layer-by-layer message passing mechanism to abstract and analyze the comparative information of two features. Fig. \ref{fig:introduction} visualizes the main differences of the proposed S-RGN, H-RGN, and other existing methods.

Our key contributions are summarized as follows:
\begin{itemize}
\item[1)] In contrast to recent works for kinship verification which mainly aim to learn discriminative features for each facial image, we have developed graph-based methods for the face matching stage to better exploit the genetic relations of two features.
\item[2)] We have presented a star-shaped reasoning graph network, where a star graph is constructed for a pair of features.
The genetic relations of two individuals are effectively exploited by analysing the incoming surrounding node messages and sending comprehensive signals to all surrounding nodes.
\item[3)] We have proposed a hierarchical reasoning graph network to further boost the reasoning capacity. Specifically, we introduce multiple latent layers to build a hierarchical graph. Then a powerful reasoning ability is obtained by hierarchically abstracting comparative information and propagating comprehensive signals on this graph.
\item[4)] We have conducted extensive experiments on four kinship verification datasets to validate the efficacy of the proposed S-RGN and H-RGN. The experimental results illustrate that our methods achieve state-of-the-art results.
\end{itemize}

It is to be noted that this paper is an extended version of our previous conference work~\cite{li2020graph}. As an extension, we have extended the star-shaped reasoning graph network into a hierarchical reasoning graph network to exploit more powerful reasoning capacity. Moreover, we have verified that our methods are not only suitable for bi-subject (one-versus-one) kinship verification, but also for tri-subject (one-versus-two) kinship learning. Furthermore, we have conducted experiments on two additional kinship verification databases to further demonstrate the efficacy of our proposed methods.  Besides, we have presented more in-depth experimental analysis and parameter discussion.

\section{Related Work}
In this section, we briefly review two related topics: 1) kinship verification, and 2) graph neural networks.

\subsection{Kinship Verification}
In the past few years, many methods~\cite{yan2014discriminative,kohli2018supervised,fang2010towards,zhou2019learning} have been proposed for kinship verification and most of them pay attention to extracting discriminative features for each facial image. We can divide them into three categories: hand-crafted approaches, distance metric-based approaches, and deep learning-based approaches.

Hand-crafted approaches require humans to design the feature extractors by hand. Traditional methods such as Principal Component Analysis (PCA), gradient orientation pyramid~\cite{zhou2012gabor}, and scale-invariant feature transform (SIFT)~\cite{somanath2012can} were frequently applied in kinship verification~\cite{liu2012tensor}. PCA was used to extract discriminative features from an ensemble of vectors. However, PCA was operated on one-dimensional vectors, which resulted in the loss of spatial information of images. Besides, as one of the earliest kinship works, Fang \emph{et al.}~\cite{fang2010towards} proposed extracting  facial parts, facial distances, color, and gradient histograms as the kinship features for verification.  A Gabor-based gradient orientation pyramid (GGOP) feature representation approach was further presented by Zhou \emph{et al.}~\cite{zhou2012gabor} to make better use of multiple feature information. Cui \emph{et al.}~\cite{cui2013fusing} proposed a face feature extracting method, which was known as the spatial face region descriptor (SFRD).

While some of the advanced hand-crafted methods are robust to the variation of illumination, rotation, and so on, they are still limited in discovering intrinsic features and coping with complicated condition variation. Distance metric-based approaches ~\cite{zhao2018learning,mahpod2018kinship} are the most popular approaches for kinship learning, which aim to learn a metric such that the distance between positive samples is reduced and that of negative face pairs is enlarged. Dehghan \emph{et al.}~\cite{dehghan2014look} proposed a model that applied gated autoencoders to extract genetic features along with metrics. Yan \emph{et al.}~\cite{yan2014discriminative} first extracted different features with multiple descriptors and then learned several distance metrics to better exploit the complementary and discriminative information.
Zhou \emph{et al.}~\cite{zhou2019learning} explicitly modeled the discrepancy of cross-generation and presented a kinship metric learning approach with a coupled deep neural network to improve the performance.
A discriminative deep metric learning approach was further introduced in ~\cite{lu2017discriminative}, which employed deep neural networks to learn a set of hierarchical nonlinear transformations.

Recent years have witnessed the extraordinary success~\cite{qiao2021efficient,fan2020deeptmm,liu2020depthpr,li2020progressivetcyb} of deep convolution neural networks. However, few deep learning-based methods~\cite{wang2020kinship,BMVC2015_148} have been proposed for kinship verification. Zhang \emph{et al.}~\cite{BMVC2015_148} presented the first deep learning-based kinship method and demonstrated the effectiveness of the proposed approach. Video-based kinship verification was further studied by Hamdi~\cite{dibeklioglu2017visual} with deep learning methods. All these methods only pay attention to learning good feature representations, which neglect how to fuse the obtained two facial image features.

Some other closely related works include~\cite{dahan2020unified,zhou2020facial,zhou2018visually}. Dahan \emph{et al.}~\cite{dahan2020unified} proposed a unified multi-task learning scheme for kinship verification which jointly learned all kinship classes. The cascaded $1 \times 1$ convolutions were used for fusion. A deep joint label distribution and metric learning (DJ-LDML) method was presented in~\cite{zhou2020facial} to exploit label relevance inherent in depression data and learn a deep ordinal embedding. Zhou \emph{et al.}~\cite{zhou2018visually} proposed the DepressNet to learn a depression representation with visual explanation and achieved excellent performance. Different from these methods, our proposed method considers how to compare and fuse two facial image features with specially designed graphs to infer the genetic relations between them.

\subsection{Graph Neural Networks}
Many kinds of applications cope with the complicated non-Euclidean structure of graph data in practice. Frasconi \emph{et al.}~\cite{frasconi1998general} first introduced a model to unify structural data such as sequences and non-structural data such as graphs. Recursive neural networks were applied to learn the transition between input graphical space and output graphical space. Graph neural networks (GNNs), which learn features on graphs, are designed to handle graph-structural data. Bruna \emph{et al.}~\cite{bruna2013spectral} proposed two methods to apply convolutional neural networks (CNN) on graph-structured data. One method was to implement convolution on the spectrum of the graph. The other method was to cluster nodes hierarchically. Henaff \emph{et al.}~\cite{henaff2015deep} studied how to construct a deep convolutional network on graph data. They proposed a parameterization method based on Spectral Networks. Gated graph neural networks (GG-NNs) were proposed by Li \emph{et al.}~\cite{li2015gated} with gated recurrent units, which was able to be trained using modern optimization techniques. Kipf and Welling proposed the graph convolutional networks (GCNs)~\cite{kipf2016semi} implementing the convolutional operation on graph-structured input, motivated by the significant performance of convolutions on the two-dimensional data. GCNs applied layer-wise propagation rule. Besides, not only node features but also local graph structure were utilized for the semi-supervised learning problem. The graph attention networks (GATs) was proposed by Petar \emph{et al.}~\cite{velivckovic2017graph} to deploy weights to a variety of nodes differently by a self-attentional method. No prior knowledge in terms of the graph structure was required to generate the weights automatically. GATs were more efficient in computation and were more capable of feature extraction thanks to the attention mechanism. Hamilton \emph{et al.}~\cite{hamilton2017inductive} proposed an inductive method to extract node information, which was known as GraphSAGE. This model used pooling across neighbor nodes when calculating embeddings of one node.

Researchers have proven that GNNs are good tools to formulate relations~\cite{li2020grapheccv}. For instance, Sun \emph{et al.}~\cite{sun2019relational} applied a recurrent graph in an action forecasting task, to formulate interactions among several objects temporally and spatially.  Gao \emph{et al.}~\cite{gao2020multi} proposed a multi-modal graph neural network, which consisted of three sub-graphs, depicting visual, semantic, and numeric modalities respectively. Then the message passing was performed to jointly reason on vision and scene text. Relational graph convolutional networks were introduced by Schlichtkrull \emph{et al.}~\cite{schlichtkrull2018modeling}, which were specifically designed to handle the highly multi-relational data characteristic of realistic knowledge bases.

\begin{figure*}[th]
  \centering
  \includegraphics[width=0.9\linewidth]{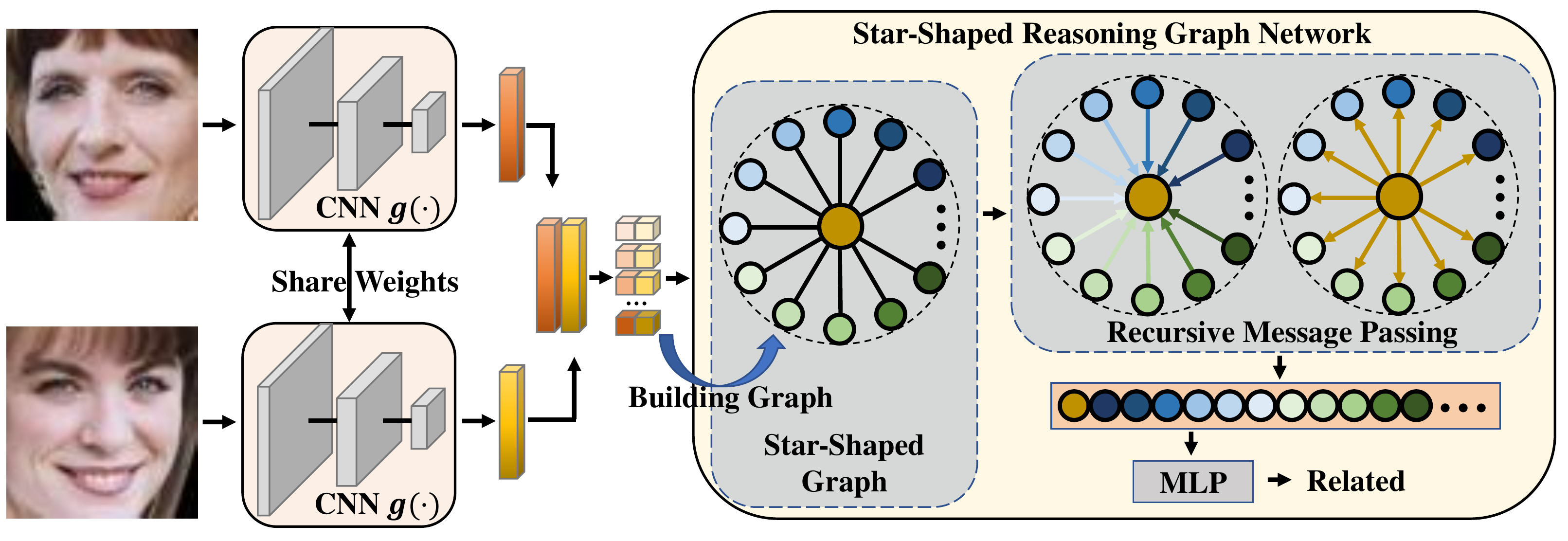}
  \caption{ An overall framework of our proposed S-RGN. To extracted image features, we first send a given image pair to the same CNN. Then we build a star-shaped reasoning graph with these two deep features and initialize each surrounding node  with the values of two deep features in one dimension. A recursive message passing scheme is employed to perform relational reasoning on the star-shaped graph. In the end, we concatenate all node features and send them to a MLP to attain the final prediction. All networks in the framework are trained end-to-end.}
  \label{fig:framework}
  \vspace{-0.5cm}
\end{figure*}

\section{The proposed methods}

In this section, we first introduce the problem formulation. Then we demonstrate the details of the proposed S-RGN and H-RGN and present how to compare and fuse two obtained facial image features.

\subsection{Problem Formulation}
We employ $\mathcal{P} = \{(\bm{x}_i,\bm{y}_i)|i=1,2,...,M\}$ to denote the training set of paired facial images with kin relations, where $\bm{x}_i$ is the parent image, $\bm{y}_i$ is the child image, and $M$ is the size of the positive training set. Therefore, the negative training set is constructed as $\mathcal{N} = \{(\bm{x}_i,\bm{y}_j)| i,j = 1,2,...,M, i \neq j\}$, where a negative sample is formed by a parent image and an unrelated child image. However, the size of the positive training set is much smaller than that of the negative training set given that $|\mathcal{P}| = M$ and $|\mathcal{N}| = M(M-1)$. So we build a balanced negative training set $\mathcal{N}'$ by randomly selecting negative samples such that $|\mathcal{N}'| = M$. Then we construct the whole training set $\mathcal{D}$  with the union of the  negative training set and positive training set:  $\mathcal{D} = \mathcal{N}' \bigcup  \mathcal{P}$.

We can formulate the goal of kinship verification as learning a mapping function, where the input is a pair of facial images $(\bm{x}_i,\bm{y}_j)$ and the output is the probability value of $i = j$. Most existing methods focus on the face representation stage and aim to learn an excellent feature extractor $g(\cdot)$. Hand-crafted methods generally design shallow image features by hand to implement the extractor $g(\cdot)$, whereas deep learning-based approaches usually learn a deep convolution neural network as the extractor $g(\cdot)$. Metric learning-based approaches usually first employ hand-crafted features or deeply learned features as the initial sample features $(g'(\bm{x}_i),g'(\bm{y}_j))$, and then learn a distance metric:
\begin{equation}
d(\bm{x}_i,\bm{y}_j) = \sqrt{d'(\bm{x}_i,\bm{y}_j)^T \bm{W} \bm{W}^T d'(\bm{x}_i,\bm{y}_j)},
\label{SRGN:metric}
\end{equation}
where $d'(\bm{x}_i,\bm{y}_j) = g'(\bm{x}_i) - g'(\bm{y}_j)$ and $\cdot^T$ denotes transposition. In the end, we attain the projected features $g(\bm{x}_i) = \bm{W}^T g'(\bm{x}_i) \in \mathbb{R}^{D}$ and $g(\bm{y}_j) = \bm{W}^T g'(\bm{y}_j) \in \mathbb{R}^{D}$, where $D$ represents the feature dimension.

Having obtained the image features $(g(\bm{x}_i),g(\bm{y}_j)) \in (\mathbb{R}^{D},\mathbb{R}^{D})$, we now need to learn a function $f(\cdot)$, which maps the features $(g(\bm{x}_i),g(\bm{y}_j))$ to a probability of kinship relation between $\bm{x}_i$ and $\bm{y}_j$. Most existing methods mainly focus on the design of feature extractors $g(\cdot)$ and usually neglect the mapping function $f(\cdot)$. One simple choice is to concatenate two image features and feed them to a multilayer perceptron (MLP):
\begin{equation}
f_{MLP}(g(\bm{x}_i),g(\bm{y}_j)) = {\rm MLP}([g(\bm{x}_i)||g(\bm{y}_j)]),
\label{SRGN:fuse_MLP}
\end{equation}
where $||$ denotes the concatenation operation. Another widely used way is to compute the cosine similarity of two image features:
\begin{equation}
f_{cos}(g(\bm{x}_i),g(\bm{y}_j)) = \frac{g(\bm{x}_i)^Tg(\bm{y}_j)}{\Vert g(\bm{x}_i)\Vert  \Vert  g(\bm{y}_j) \Vert  }.
\label{SRGN:fuse_cos}
\end{equation}
While the image features usually contain rich semantic information, the MLP solution simply concatenates two features and fails to explicitly model the corresponding relations between two features. The cosine solution only considers the angle between two feature vectors and ignores the rich semantic information encoded in the image features. Therefore, both methods cannot effectively exploit the relations of two image features. In this work, we aim to design a new mapping $f(\cdot)$ to perform relational reasoning on the two features.

\subsection{Building a Star-Shaped Reasoning Graph Network}
Recently, deep convolutional neural networks have proven to be effective in many computer vision problems, including image classification~\cite{he2016deep}, scene understanding~\cite{yang2020learning}, and object detection~\cite{faster17pami}, which exhibits their outstanding capacity in feature representation. Therefore, a deep CNN which is parameterized by $\bm{\Omega}$ is applied as the feature extractor $g(\cdot,\bm{\Omega})$ in this paper.

Having attained the deep features $(g(\bm{x}_i,\bm{\Omega}),g(\bm{y}_j,\bm{\Omega}))$, we consider performing relation reasoning on these features. Relation reasoning can be achieved by observing how humans reason about kin relations. As facial characteristics usually exhibit the genetic traits, humans may predict the kinship relations comparing the related attributes on two faces to discover the hidden genetic similarity. For instance, if we observe that there are the same eye color and similar cheekbones on the two persons' facial images, the probability will be higher of them being related. After comparing several genetic facial features of two persons, humans analyze and combine this information to make the decision finally.

We construct a graph to model such a reasoning process explicitly. Relational reasoning is then performed on this graph. We assume that various genetic information is encoded on each dimension of the extracted features. Kinship relations can be reasoned by comparing and fusing this information. Because the same CNN is applied to extract features from two images, the values in the same dimension of two features represent the same type of kinship related genetic information in the corresponding dimension. Each feature dimension occupies one node in the graph as a comparison, then there would be $D$ nodes describing the comparisons of all dimensions in two visual features. To fuse such comparison information, the interactions of these nodes should be defined. An intuitive method is to establish connections between all the nodes considering that any two nodes may be related. However, a graph using such a connecting method would result in greatly increased computational complexity. To address this problem, a latent node is introduced which is connected to other nodes. Further, the nodes are not connected to each other but only connected to the latent node. As the center of the star-structured graph, the latent node plays a significant role in the communication and interaction of information among $D$ surrounding nodes.

After building the star-shaped graph, performing relational reasoning on the graph should be formulated. In recent years, graph neural networks (GNNs) have gained more and more attention~\cite{valsesia2020deep,ji2020context} in representation learning of graph-structural data. In a nutshell, GNNs apply the recursive message-passing scheme, where all nodes aggregate the messages from its neighbors and update its own feature. To reason on the star-shaped graph, we apply such a scheme and propose the star-structural reasoning network.

Specifically, denote $\mathcal{G}_{S} = (\mathcal{V}_{S},\mathcal{E}_{S})$ as the graph containing the node-set $\mathcal{V}_{S}$ along with the edge set $\mathcal{E}_{S}$. Nodes in the graph are represented with feature vectors, so $\mathcal{V}_{S} = \{\bm{h}_c \} \bigcup \{\bm{h}_d|d=1,2,...,D\}$, where $\bm{h}_c$ indicates the feature vector of the latent node, at the same time, $\bm{h}_d$ is the feature vector of $d^{th}$ peripheral node. The edge set of this graph can be formulated as $\mathcal{E}_{S} = \{e_{cd} | d = 1,2,...,D\}$, which means $e_{cd}$ indicates the edge between node $\bm{h}_c$ and $\bm{h}_d$. The proposed S-RGN forwards messages based on the graph structure as known as $\mathcal{E}$. Messages are aggregated to update the features of nodes. As described above, we put the values of the same dimension from two  extracted image features as the initial features of the surrounding node. Formally, the initial features of the surrounding node are set as:
\begin{equation}
\bm{h}_d^0 = [g_d(\bm{x}_i,\bm{\Omega}) || g_d(\bm{y}_j,\bm{\Omega})],
\label{SRGN:ini_node}
\end{equation}
where $\bm{h}_d^0 \in \mathbb{R}^{2}$ indicates the $d^{th}$ node's initial feature, $g_d(\bm{x}_i,\bm{\Omega})$ and $g_d(\bm{y}_j,\bm{\Omega})$ denodes the values of the $d^{th}$ dimension from features  $g(\bm{x}_i,\bm{\Omega})$ and $g(\bm{y}_j,\bm{\Omega})$. By this means, each surrounding node encodes one specific genetic kinship feature. Subsequently, to initialize the feature of center node, we propose to utilize the features of all other nodes $\{\bm{h}_1^{0},\bm{h}_2^{0},...,\bm{h_}D^{0} \}$ considering the center node is related to all other nodes:
\begin{equation}
\bm{h}_c^0 = INIT_S(\bm{h}_1^{0},\bm{h}_2^{0},...,\bm{h_}D^{0}),
\label{SRGN:ini_center_node}
\end{equation}
where the $INIT_S(\cdot)$ indicates a mapping function which can be formulated as pooling function.

\begin{algorithm}[t]
	\DontPrintSemicolon
	\KwIn{Training set: $\mathcal{D}$, Parameters: $\Gamma$ (iterative number), $K$ (layer number of the S-RGN), and  \{$F_1$, $F_2$, $...$, $F_K$\} (feature dimensions of $K$ layers).\;
	}
	\KwOut{The weights $\bm{\Omega}$ of the feature extractor $g(\cdot,\bm{\Omega})$, and the weights of the S-RGN $\bm{\theta}_{S}$ = $\{ \bm{W}_{mess,1}$, $\bm{W}_{surr,1}$, $\bm{W}_{cen,1}$, $..., \bm{W}_{mess,K}$, $\bm{W}_{surr,K}$, $\bm{W}_{cen,K}$, $\bm{\Theta}_{S}\}$.}
	\SetAlgoLined	
    Initialize parameters $\bm{\Omega}$ and $\bm{\theta}_{S}$.\;
	\For{$iter \leftarrow$ 1, 2, ..., $\Gamma$}
     {
       Sample a mini-batch from the training set.\;
       Extract deep features with the CNN $g(\cdot,\bm{\Omega})$.\;
       Build the initial star-shaped reasoning graph with \eqref{SRGN:ini_node} and \eqref{SRGN:ini_center_node}.\;
       \For{$k \leftarrow$ 1, 2, ..., $K$}
       {
         Generate the message of all nodes for $k^{th}$ layer using \eqref{SRGN:mess1} and \eqref{SRGN:mess2}.\;
         Update the surrounding node features with \eqref{SRGN:update1}.\;
         Update the central node feature with \eqref{SRGN:agg} and \eqref{SRGN:update2}.\;
       }
       Concatenate the final features of all nodes.\;
       Send the concatenated features to a MLP to obtain the predictions with \eqref{SRGN:res}.\;
       Compute the loss $\mathcal{L}_{S}$ with \eqref{SRGN:loss}.\;
       Update the parameters $\bm{\Omega}$ and $\bm{\theta}_{S}$ by descending the stochastic gradient: $\nabla \mathcal{L}_{S}$.
    }
\textbf{Return:} The parameters $\bm{\Omega}$ and $\bm{\theta}_{S}$.
\caption{ The training procedure of our S-RGN} \label{alg:train_SRGN}	
\end{algorithm}

\subsection{Reasoning with the S-RGN}

Having obtained the initial graph, we consider how to perform relational reasoning with recursive message passing. The proposed S-RGN has $K$ layers where each layer stands for one time-step of the message passing phase. The $k^{th} (1 \leq k \leq K)$ layer is responsible for transforming the node features $\bm{h_}c^{k-1},\bm{h}_1^{k-1},\bm{h}_2^{k-1},...,\bm{h}_D^{k-1} \in \mathbb{R}^{F_{k-1}}$ into $\bm{h}_c^{k},\bm{h}_1^{k},\bm{h}_2^{k},...,\bm{h}_D^{k} \in \mathbb{R}^{F_{k}}$ with message passing, where $\mathbb{R}^{F_{k-1}}$ and $\mathbb{R}^{F_{k}}$ denote the corresponding feature dimensions. Assuming that we have attained the node features of the $(k-1)^{th}$ layer, we first generate the node messages which are going to be sent out in the following message passing step. We generate the messages of surrounding nodes as follows:
\begin{equation}
\bm{m}_d^{k} = {\rm ReLU}(\bm{W}_{mess,k}^T \bm{h}_d^{k-1}), d= 1,2,...,D
\label{SRGN:mess1}
\end{equation}
where $\bm{W}_{mess,k} \in \mathbb{R}^{F_{k-1} \times F_{k}}$ is utilized to transform the node features into node messages in the $k^{th}$ layer. The same operation is applied for the central node:
\begin{equation}
\bm{m}_c^{k} = {\rm ReLU}(\bm{W}_{mess,k}^T \bm{h}_c^{k-1}).
\label{SRGN:mess2}
\end{equation}

With these generated messages, we propagate and aggregate them based on the graph structure defined by $\mathcal{E}_{S}$. Then the node features are updated with the aggregated messages. For the surrounding nodes, given that the central node is the only neighborhood, we aggregate node messages by concatenating the message of the central node and its own message. Then the aggregated messages are used to update the node feature following:
\begin{equation}
\bm{h}_d^{k} = {\rm ReLU}(\bm{W}_{surr,k}^T [\bm{m}_d^{k} || \bm{m}_c^{k} ] ), d= 1,2,...,D
\label{SRGN:update1}
\end{equation}
where $\bm{W}_{surr,k} \in \mathbb{R}^{2F_{k} \times F_{k}}$ is employed to fuse all information to attain the updated node feature. For the central node, all the incoming messages are first aggregated:
\begin{equation}
\bm{m}_a^{k} = AGGRE_S(\{ \bm{m}_d^k | d = 1,2,...,D\}),
\label{SRGN:agg}
\end{equation}
where $AGGRE_S(\cdot)$ is an aggregate function, which is implemented by a pooling operation. Then we update the feature of the central node as follows:
\begin{equation}
\bm{h}_c^{k} = {\rm ReLU}(\bm{W}_{cen,k}^T [\bm{m}_c^{k} || \bm{m}_a^{k} ] ),
\label{SRGN:update2}
\end{equation}
where $\bm{W}_{cen,k} \in \mathbb{R}^{2F_{k} \times F_{k}}$ is used to update the feature of the central node. In this way, we attain the updated node features $\bm{h}_c^{k},\bm{h}_1^{k},\bm{h}_2^{k},...,\bm{h}_D^{k}$ by message passing.

We iterate the above process for $K$ times and obtain the final node feature vectors: $\bm{h}_c^{K},\bm{h}_1^{K},\bm{h}_2^{K},...,\bm{h}_D^{K} \in \mathbb{R}^{F_{K}}$. To make the final prediction, we first concatenate all these features and feed them to a MLP $\psi_S(\cdot,\bm{\Theta}_S)$, which outputs a scalar value. Therefore, we can formulate the mapping function $f_S(\cdot)$ of our S-RGN as:
\begin{equation}
f_S(g(\bm{x}_i),g(\bm{y}_j)) = \psi_S([\bm{h}_c^{K} || \bm{h}_1^{K} || \bm{h}_2^{K} || ... || \bm{h}_D^{K} ],\bm{\Theta}_S),
\label{SRGN:res}
\end{equation}
where $\bm{\Theta}_S$ is the learnable parameters of MLP. Lastly, we apply a sigmoid function $\sigma(\cdot)$ to the value $f_S(g(\bm{x}_i),g(\bm{y}_j))$ and obtain the probability value of kin relation between $\bm{x}_i$ and $\bm{y}_j$. Fig. \ref{fig:framework} depicts the above pipeline.

It should be noted that our S-RGN and the feature extractor $g(\cdot)$ are trained end-to-end.  The binary cross-entropy loss is employed as the objective function:
\begin{equation}
\begin{aligned}
\mathcal{L}_S =& - \frac{1}{N} \sum_{(\bm{x},\bm{y}) \in \mathcal{P}} \log(\sigma( f_S(g(\bm{x}),g(\bm{y}))))  \\
 & - \frac{1}{N} \sum_{(\bm{x},\bm{y}) \in \mathcal{N}'}  \log(1 - \sigma (f_S(g(\bm{x}),g(\bm{y})))).
\end{aligned}
\label{SRGN:loss}
\end{equation}

In this way, the proposed S-RGN is optimized in a class-balanced setting. Lastly, we summarize the training procedure of the proposed S-RGN in Algorithm \ref{alg:train_SRGN}.

\subsection{Building a Hierarchical Reasoning Graph Network}

To reason with $D$ visual comparison nodes, the S-RGN introduce a latent reasoning node to interact with these $D$ nodes. Although the constructed star graph effectively reduces the computational complexity of information interaction between $D$ visual comparison nodes, it also limits the reasoning ability and flexibility of our model. Instead of employing a latent reasoning node, we first consider introducing a latent layer with a set of latent nodes to address this issue. Consequently, the information of visual comparison nodes is propagated and abstracted through a shared latent space of multiple latent reasoning nodes.
Furthermore, we consider that when humans infer the kin relation of two individuals, humans do not simultaneously process all visual contrast information, but often organize and analyze the information in a local and hierarchical way. That is to say, people may first  locally combine some fine level of comparative information to synthesize the middle-level preferences, and then analyse these middle-level preferences to make the final judgment. Therefore, we propose to infer and aggregate information in a hierarchical way by introducing multiple latent reasoning layers to exploit more powerful reasoning ability. Then, for any two adjacent latent reasoning layers, the upper layer is responsible for aggregating the information from the lower layer and propagating the received comprehensive signals to the lower layer.

Formally, Let $\mathcal{G}_{H} = (\mathcal{V}_{H},\mathcal{E}_{H})$ denote the hierarchical latent reasoning graph with the node set $\mathcal{V}_{H}$ and the edge set $\mathcal{E}_{H}$.  We assume that the graph $\mathcal{G}_{H}$ has $L$ latent reasoning layers, and for the $l^{th} (1 \leq l \leq L)$ latent layer, the number of nodes is $N_l$. If we regard the $D$ visual comparison nodes as the $0^{th}$ layer of the graph $\mathcal{V}_{H}$ where $N_0 = D$, then we have $\mathcal{V}_{H} = $ $\{ \bm{h}_{0,1}, ..., \bm{h}_{0,N_0}, ..., \bm{h}_{l,1}, ..., \bm{h}_{l,n_l}, ..., \bm{h}_{l,N_l}, ..., \bm{h}_{L,1}, ..., \bm{h}_{L,N_L} \}$, where $\bm{h}_{l,n_l}$ represents the $n_l^{th} (1 \leq n_l \leq N_l)$ reasoning node of the $l^{th}$ latent layer, and generally $N_0 > N_1 > ... > N_L$. As the hierarchy of the human reasoning process often cooperates with the locality, we do not simply connect all the nodes between adjacent two latent layers. Instead, we consider a sparse, locally connected and tree-like graph structure, where for two adjacent latent reasoning layers, a node in the lower layer is only connected with a node in the upper layer, while a node in the upper layer is connected with several consecutive nodes in the lower layer. Mathematically, we use a set of adjacency matrices $\bm{A}$ to describe the topology of our hierarchical graph: $\bm{A} = \{\bm{A}_{0,1}, ..., \bm{A}_{l-1,l}, ..., \bm{A}_{L-1,L} \} $, where $\bm{A}_{l-1,l} \in \{0,1\}^{N_{l-1} \times N_l} $ denotes the adjacency matrix between the latent layer $l-1$ and the latent layer $l$.  The element $\bm{A}_{l-1,l}^{n_{l-1},n_l}$ denotes the connectivity between the $n_{l-1}^{th} (1 \leq n_{l-1} \leq N_{l-1})$ node of layer $l-1$ and the $n_l^{th} (1 \leq n_l \leq N_l)$ node of layer $l$. We set $C_{l-1,l} = (N_{l-1} \mod N_{l})$.
When  $n_l \leq C_{l-1,l}$, we have
\begin{equation}
\bm{A}_{l-1,l}^{n_{l-1},n_l}=
\begin{cases}
1& \lceil \frac{N_{l-1}}{N_l} \rceil (n_l -1) < n_{l-1} \leq \lceil \frac{N_{l-1}}{N_l} \rceil n_l\\
0& \text{others}
\end{cases}
\label{HLRG:adja1}
\end{equation}
When  $n_l > C_{l-1,l}$, we have
\begin{equation}
\bm{A}_{l-1,l}^{n_{l-1},n_l}=
\begin{cases}
1& \lfloor \frac{N_{l-1}}{N_l} \rfloor (n_l - 1)  < n_{l-1} - C_{l-1,l} \leq \lfloor \frac{N_{l-1}}{N_l} \rfloor n_l\\
0& \text{others}
\end{cases}
\label{HLRG:adja2}
\end{equation}

Having obtained the graph structure defined by $\bm{A}$, now we consider how to set the initial features of the nodes $\mathcal{V}_{H}$. For $D$ visual comparison nodes, we use the same initial features as the S-RGN method:
\begin{equation}
\bm{h}_{0,n_0}^0 = [g_{n_0}(\bm{x}_i,\bm{\Omega}) || g_{n_0}(\bm{y}_j,\bm{\Omega})], (1 \leq n_0 \leq D).
\label{HLRG:ini_node}
\end{equation}

We consider initializing the upper latent reasoning layers with the initial node features of the lower layers. More specifically, for the $n_l^{th}$ node in the $l^{th}$ layer, we initialize it as follows:
\begin{equation}
\bm{h}_{l,n_l}^0 = INIT_H(\{ \bm{h}_{l-1,s}^0 | s \in \mathcal{S}_{l,n_l}\}),
\label{HLRG:ini_hie_node}
\end{equation}
where $\mathcal{S}_{l,n_l} = \{s | \bm{A}_{l-1,l}^{s,n_l} = 1 \} $ and $INIT_H(\cdot)$ is a mapping function. Unlike the initialization function $INIT_S(\cdot)$ in the S-RGN method, which uses a simple pooling operation, we adopt a self-attention mechanism for $INIT_H(\cdot)$ to extract more discriminative initial features. Concretely, we apply a MLP $a(\cdot)$, which is parameterized by $\bm{\Psi}$ to compute attention coefficients:
\begin{equation}
\alpha_s = a(\bm{h}_{l-1,s}^0,\bm{\Psi}), s \in \mathcal{S}_{l,n_l},
\label{HLRG:attention}
\end{equation}
which indicate the importance of node $\bm{h}_{l-1,s}$ to node $\bm{h}_{l,n_l}$ in the initialization phase. Then we initialize the node $\bm{h}_{l,n_l}$ with normalized self-attention coefficients:
\begin{equation}
\bm{h}_{l,n_l}^0 = \sum_{s \in \mathcal{S}_{l,n_l}} \frac{\exp(\alpha_s)}{ \sum_{s' \in \mathcal{S}_{l,n_l}}\exp(\alpha_{s'}) } \bm{h}_{l-1,s}^0.
\label{HLRG:ini_hie_node2}
\end{equation}
In this way, we initialize our hierarchical latent reasoning graph layer by layer.

\subsection{Reasoning with the H-RGN}

Having obtained the initialized hierarchical graph, we consider the message passing mechanism of the proposed H-RGN. Just like our S-RGN, we also stack $K$ layers of H-RGN, where each H-RGN layer is responsible for a complete step of message passing. Therefore, the $k^{th}$ H-RGN layer transforms the node features  $\bm{h}_{0,1}^{k-1}, ..., \bm{h}_{0,N_0}^{k-1}, ...,$ $\bm{h}_{l,1}^{k-1}, ..., \bm{h}_{l,N_l}^{k-1}, ...,$ $\bm{h}_{L,1}^{k-1}, ..., \bm{h}_{L,N_L}^{k-1}$ $\in \mathbb{R}^{F_{k-1}}$ into $\bm{h}_{0,1}^{k}, ..., \bm{h}_{0,N_0}^{k},$ $..., \bm{h}_{l,1}^{k}, ..., \bm{h}_{l,N_l}^{k},$ $..., \bm{h}_{L,1}^{k}, ..., \bm{h}_{L,N_L}^{k}$ $\in \mathbb{R}^{F_{k}}$, where $F_{k-1}$ and $F_{k}$ represent the updated feature dimensions of nodes in the H-RGN layer $k-1$ and $k$, respectively. We elaborate on the message passing process at the $k^{th}$ H-RGN layer to show how the proposed method performs relational reasoning on the hierarchical graph.

The $k^{th}$ H-RGN layer first transforms the node features of $(k-1)^{th}$ layer to obtain the transformed node features:
\begin{equation}
\bm{m}_{l,n_l}^{k} = {\rm ReLU}(\bm{U}_{trans,k}^T \bm{h}_{l,n_l}^{k-1}), 0 \leq l \leq L, 1 \leq n_l \leq N_l,
\label{HLRG:trans}
\end{equation}
where $\bm{U}_{trans,k} \in \mathbb{R}^{F_{k-1} \times F_{k}}$ is a learnable weight matrix.
Considering that each H-RGN layer has $L$ latent reasoning layers, a complete message passing step on the $L$-layer hierarchical graph includes a bottom-up comparative information abstraction stage and a top-down comprehensive signal propagation stage. The bottom-up stage aims at aggregating and abstracting comparative information hierarchically to obtain the comprehensive node features $\bm{c}_{0,1}^{k}, ..., \bm{c}_{0,N_0}^{k},$ $..., \bm{c}_{l,1}^{k}, ..., \bm{c}_{l,N_l}^{k},$ $..., \bm{c}_{L,1}^{k}, ..., \bm{c}_{L,N_L}^{k}$. To this end, we first set $\bm{c}_{0,n_0}^{k} = \bm{m}_{0,n_0}^{k} (1 \leq n_0 \leq N_0)$ and then use the node features of lower latent layers to update that of upper latent layers according to the graph structure:
\begin{equation}
\bm{c}_{l,n_l}^{k} = {\rm ReLU}(\bm{U}_{up,k}^T [\bm{m}_{l,n_l}^{k} || AGGRE_H(\{ \bm{m}_{l-1,s}^{k}  | s \in S_{l,n_l} \})] ),
\label{HLRG:update1}
\end{equation}
where $AGGRE_H(\cdot)$ is implemented by a pooling operation and $\bm{U}_{up,k} \in \mathbb{R}^{2F_{k} \times F_{k}}$ fuses all incoming messages. We apply \eqref{HLRG:update1} layer by layer to get the comprehensive node features: $\bm{c}_{0,1}^{k}, ..., \bm{c}_{0,N_0}^{k},$ $..., \bm{c}_{l,1}^{k}, ..., \bm{c}_{l,N_l}^{k},$ $..., \bm{c}_{L,1}^{k}, ..., \bm{c}_{L,N_L}^{k}$.

The top-down stage is devoted to propagating the comprehensive signal encoded in the comprehensive node features to the lower layers. Formally, if the node number of the top latent layer is one ($N_L = 1$), we directly set $\bm{h}_{L,1}^k$ to $\bm{c}_{L,1}^k$. Otherwise, we attain the node features of the top latent layer by propagating  signals among them:
\begin{equation}
\bm{h}_{L,n_L}^{k} = \sum_{1 \leq s \leq N_L} \phi(\bm{c}_{L,n_L}^{k},\bm{c}_{L,s}^{k}) \bm{c}_{L,s}^{k}, 1 \leq n_L \leq N_L,
\label{HLRG:top}
\end{equation}
where $\phi(\cdot,\cdot)$ represents the pairwise relations between two latent nodes. We use the cosine similarity to implement $\phi(\cdot,\cdot)$.
Subsequently, we propagate the comprehensive signal to the lower layer with the updated upper layer node features:
\begin{equation}
\bm{h}_{l,n_l}^{k} = {\rm ReLU}(\bm{U}_{down,k}^T [\bm{c}_{l,n_l}^{k} || \bm{h}_{l+1,F}^{k}] ),
\label{HLRG:update2}
\end{equation}
where $F$ denotes the node on the $(l+1)^{th}$ latent layer that satisfies $A_{l,l+1}^{n_l,F} =1$, and $\bm{U}_{down,k}  \in \mathbb{R}^{2F_{k} \times F_{k}}$ is employed to update the node features.

We obtain the set of final node features $\bm{H}^{K} = \{ \bm{h}_{0,1}^{K},$ $..., \bm{h}_{0,N_0}^{K}, ...,$ $\bm{h}_{l,1}^{K}, ..., \bm{h}_{l,N_l}^{K}, ...,$ $\bm{h}_{L,1}^{K}, ..., \bm{h}_{L,N_L}^{K} \}$ after iterating the bottom-up and top-down message passing processes for $K$ steps. Then the mapping function $f_H(\cdot)$ of the H-RGN method is implemented by concatenating these features and sending them to a MLP $\psi_H(\cdot,\bm{\Theta}_H)$ parameterized by $\bm{\Theta}_H$:
\begin{equation}
f_H(g(\bm{x}_i),g(\bm{y}_j)) = \psi_H( CONCAT(\bm{H}^{K}),\bm{\Theta}_H),
\label{HLRG:res}
\end{equation}
To obtain the predicted probability, we further normalize $f_H(g(\bm{x}_i),g(\bm{y}_j))$ with a sigmoid function $\sigma(\cdot)$. We also adopt the binary cross-entropy loss to train our H-RGN networks:
\begin{equation}
\begin{aligned}
\mathcal{L}_H =& - \frac{1}{N} \sum_{(\bm{x},\bm{y}) \in \mathcal{P}} \log(\sigma( f_H(g(\bm{x}),g(\bm{y}))))  \\
 & - \frac{1}{N} \sum_{(\bm{x},\bm{y}) \in \mathcal{N}'}  \log(1 - \sigma (f_H(g(\bm{x}),g(\bm{y})))).
\end{aligned}
\label{HLRG:loss}
\end{equation}

Lastly, we show the training procedure of the proposed H-RGN in Algorithm \ref{alg:train_HRGN}.

\begin{algorithm}[t]
	\DontPrintSemicolon
	\KwIn{Training set: $\mathcal{D}$, Parameters: $\Gamma$ (iterative number), $K$ (layer number of the H-RGN), \{$F_1$, $F_2$, $...$, $F_K$\} (feature dimensions of $K$ H-RGN layers), and \{$N_1$, $N_2$, ..., $N_L$\} (node numbers of $L$ latent layers) \;
	}
	\KwOut{The weights $\bm{\Omega}$ of the feature extractor $g(\cdot,\bm{\Omega})$, and the weights of the H-RGN $\bm{\theta}_{H}$ = $\{\bm{\Psi}$, $\bm{U}_{trans,1}$, $\bm{U}_{up,1}$, $\bm{U}_{down,1}$, $..., \bm{U}_{trans,K}$, $\bm{U}_{up,K}$, $\bm{U}_{down,K}$, $\bm{\Theta}_{H}\}$.}
	\SetAlgoLined	
    Initialize parameters $\bm{\Omega}$ and $\bm{\theta}_{H}$.\;
	\For{$iter \leftarrow$ 1, 2, ..., $\Gamma$}
     {
       Sample a mini-batch from the training set.\;
       Extract deep features with the CNN $g(\cdot,\bm{\Omega})$.\;
       Build the initial hierarchical reasoning graph with \eqref{HLRG:ini_node} - \eqref{HLRG:ini_hie_node2}.\;
       \For{$k \leftarrow$ 1, 2, ..., $K$}
       {
         Transform node features with \eqref{HLRG:trans}.\;
         // \emph{bottom-up comparative information abstraction}.\;
         Set $\bm{c}_{0,n_0}^{k} = \bm{m}_{0,n_0}^{k} (1 \leq n_0 \leq N_0)$.\;
         Aggregate comprehensive node features layer by layer using \eqref{HLRG:update1}.\;
         // \emph{top-down comprehensive signal propagation}.\;
         \eIf{$N_L = 1$}
         {
           Set $\bm{h}_{L,1}^k  = \bm{c}_{L,1}^k$.\;
         }
         {
           Propagate comprehensive signals among the top layer nodes with \eqref{HLRG:top}.\;
         }
         Update node features hierarchically down to the bottom latent layer using \eqref{HLRG:update2}.\;

       }
       Calculate the predictions with the concatenated features using \eqref{HLRG:res}.\;
       Compute the loss $\mathcal{L}_{H}$ with \eqref{HLRG:loss}.\;
       Update the parameters $\bm{\Omega}$ and $\bm{\theta}_{H}$ by descending the stochastic gradient: $\nabla \mathcal{L}_{H}$.
    }
\textbf{Return:} The parameters $\bm{\Omega}$ and $\bm{\theta}_{H}$.
\caption{The training procedure of our H-RGN} \label{alg:train_HRGN}	
\end{algorithm}

\section{EXPERIMENTS}
In this section, we conducted extensive experiments on four widely-used kinship  databases to demonstrate the effectiveness of the proposed S-RGN and H-RGN methods. The experimental results and analysis are described in detail as follows.

\begin{figure}[t]
  \centering
  \includegraphics[width=0.9\linewidth]{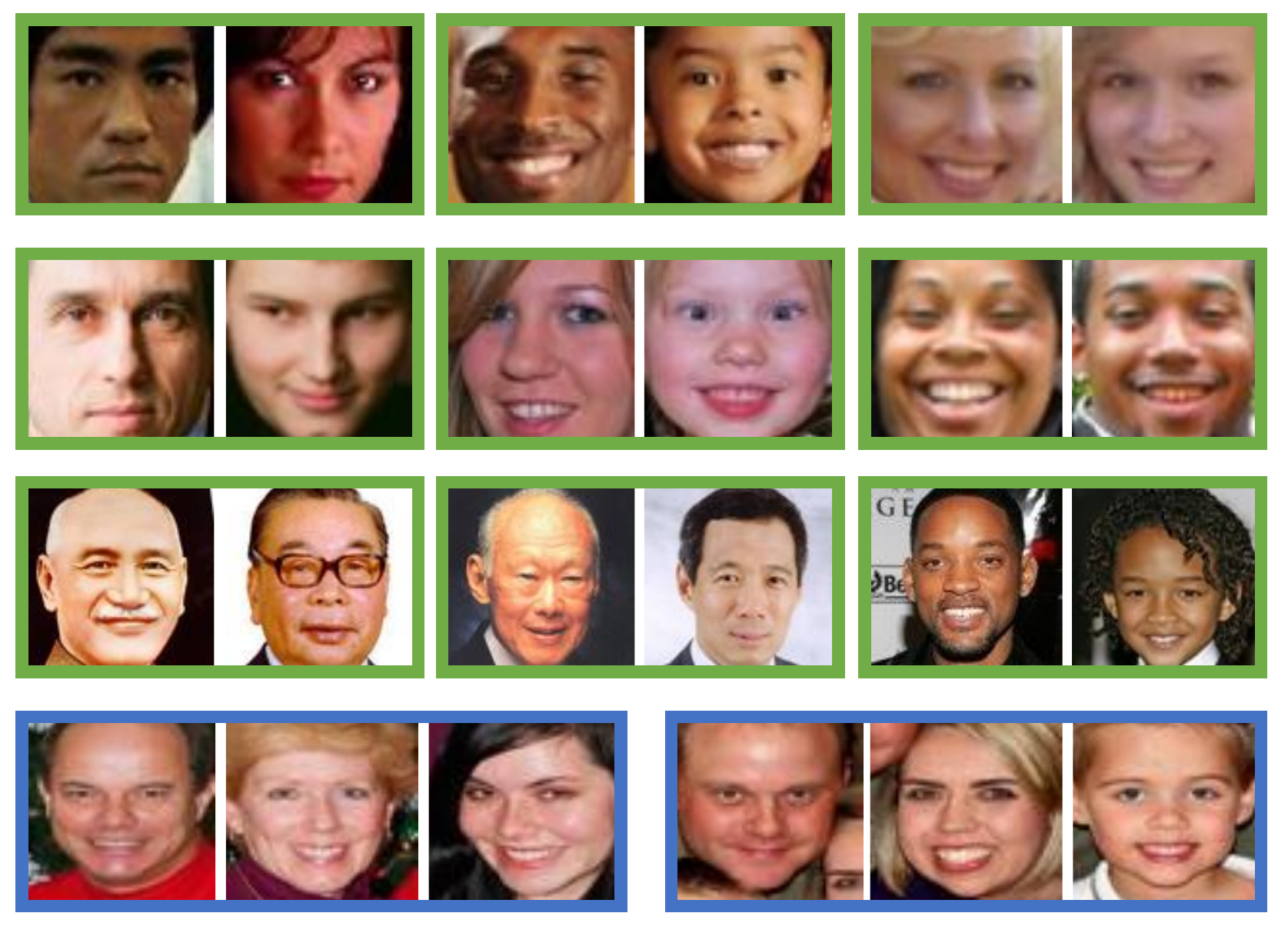}
  \caption{Some image examples positive pairs (with kinship relation) from four kinship datasets. From top to down are images from the KinFaceW-I, KinFaceW-II, Cornell KinFace, and TSKinFace datasets, accordingly. The first three rows show bi-subject kinship relations, while the last row shows tri-subject kinship relations: Father-Mother-Daughter (FM-D) and Father-Mother-Son (FM-S).}
  \label{fig:dataset}
  \vspace{-0.5cm}
\end{figure}

\subsection{Datasets and Experiment Settings}
\emph{KinFaceW-I~\cite{lu2013neighborhood} and KinFaceW-II~\cite{lu2013neighborhood} Datasets:} The KinFaceW-I and KinFaceW-II datasets are two widely-used databases for evaluation, which are collected from the internet. These two databases investigate four different types of kin relationships: Father-Son (F-S), Father-Daughter (F-D),  Mother-Son (M-S), and Mother-Daughter (M-D).
For these four relations, the KinFaceW-I dataset contains 134, 156, 127, and 116  pairs of parent-child facial images respectively whereas the KinFaceW-II database consists of 250 pairs of  facial images for each kinship relation. The key difference between these two datasets is that each parent-child image pair with kin relation in the KinFaceW-I database comes from different photos while that in the KinFaceW-II database is collected from the same photo. Each facial image is aligned and cropped of size $64 \times 64$. We adopt the five-fold cross-validation in the experiments following the standard protocol in~\cite{lu2013neighborhood}.

\emph{TSKinFace Dataset~\cite{qin2015tri}:} The TSKinFace database is constructed to investigate the tri-subject kinship verification, where the images are harvested from the internet. No restrictions such as race, lighting, and background are imposed during the collecting stage. TSKinFace dataset contains two kinds of family-based kinship relations: Father-Mother-Son (FM-S) and Father-Mother-Daughter (FM-D). The FM-S and FM-D have 513 and 502 tri-subject relations, respectively. The face images are detected and cropped to $64 \times 64$ pixels according to the eye coordinates. Each tri-subject relation is further divided into two bi-subject relations so that we report the results of six types of relations: 513 F-S, 502 F-D, 513 M-S, 502 M-D, 513 FM-S, and 502 FM-D groups.  Following the commonly used protocol as adopted in many previous works~\cite{qin2015tri,liang2018weighted}, we split the samples of each kind of relation into five groups where each group contains nearly the same number of groups and perform five-fold cross-validation.

\emph{Cornell KinFace Dataset~\cite{fang2010towards}:} The Cornell KinFace dataset consists of 150 pairs of parent-child images, which is collected through an on-line search. The face images are chosen to be frontal and a neutral facial expression to ensure image quality. The database includes 40\% father-son pairs, 22\% father-daughter pairs, 13\% mother-son pairs, and 26\% mother-daughter pairs. Due to privacy issues, 7 families are removed from the original dataset and the remaining 143 pairs of kinship images are used for validation~\cite{yan2014discriminative}. The five-fold cross-validation is conducted for each relation respectively.

Fig. \ref{fig:dataset} shows some example images of these datasets, respectively. In the experiments, the ResNet-18 was employed as the feature extractor network $g(\cdot)$, which was initialized with the weights pre-trained on ImageNet. Naturally, the feature dimension $D$  was equal to 512. Data augmentation is a crucial step to improve performance. Specifically, random cropping and flipping were utilized to augment data. Note that we train our reasoning graph network and the feature extractor network end-to-end. We used PyTorch~\cite{paszke2017automatic} to implement our algorithm and tested our methods on a system with Intel(R) Xeon(R) CPU E5-2620 v4 @ 2.10GHz. Besides, one GeForce RTX 2080 Ti GPU was employed for neural network acceleration.

\begin{table}[t]
\caption{Mean verification rate (\%) of the S-RGN method using different pooling operations for $INIT_S(\cdot)$. }
\label{table:SRGN_INIT}
\centering
\begin{tabular}{|l|c|ccccc|}
\hline
Dataset & $INIT_S(\cdot)$ & F-S & F-D & M-S & M-D & Mean \\
\hline \hline
\multirow{2}*{KinFaceW-I} & Max &  72.4 &  73.2  & 72.9 & 79.5 & 74.5 \\
 ~ & Avg & \textbf{78.8} &  \textbf{75.4}  & \textbf{80.1}  & \textbf{83.8} & \textbf{79.5} \\
\hline
\multirow{2}*{KinFaceW-II} & Max & 83.6 & 79.4 & 82.6 & 88.4 & 83.5 \\
~ & Avg & \textbf{90.8} & \textbf{87.0} & \textbf{91.0} & \textbf{93.6} & \textbf{90.6} \\
\hline
\end{tabular}
\vspace{-0.5cm}
\end{table}

To validate the effectiveness of the proposed methods, we report the performance of the following two baselines, which adopt different face matching mechanisms:
\begin{itemize}

\item MLP: This method first concatenates two extracted features and then sends them to a MLP to obtain the probability of kinship relation between two individuals. The formulation is shown in \eqref{SRGN:fuse_MLP}.
\item Cos: We use the same CNN backbone as our proposed methods to extract image features. Then the cosine similarity of two extracted features is adopted as presented in \eqref{SRGN:fuse_cos} to measure the kinship relations.
\end{itemize}

\subsection{Experimental Results on KinFaceW-I and KinFaceW-II}

\begin{table}[t]
\caption{Mean verification rate (\%) of the H-RGN method with different design choices for $INIT_H(\cdot)$. }
\label{table:HLRG_INIT}
\centering
\begin{tabular}{|l|c|ccccc|}
\hline
Dataset & $INIT_H(\cdot)$ & F-S & F-D & M-S & M-D & Mean \\
\hline \hline
\multirow{3}*{KinFaceW-I} & Max Pooling&  76.9 &  72.7  & 77.1 & 87.0 & 78.4 \\
 ~ & Avg Pooling & 80.4 &  74.4  & 76.7 & 87.4 & 79.7 \\
 ~ & Self-Attention & \textbf{81.1} &  \textbf{78.4}  & \textbf{80.2} & \textbf{87.8} & \textbf{81.9} \\
\hline
\multirow{3}*{KinFaceW-II} & Max Pooling& 88.4 & 84.2 & 91.4 & 93.2 & 89.3 \\
~ & Avg Pooling& 90.0 & \textbf{86.4} & 91.6 & 94.8 & 90.7 \\
~ & Self-Atention & \textbf{91.0} & 85.6 & \textbf{93.0} & \textbf{96.0} & \textbf{91.4} \\ 
\hline
\end{tabular}
\end{table}

\begin{table}[t]
\caption{Mean verification rate (\%) of the S-RGN method using different pooling operations for $AGGRE_S(\cdot)$.}
\label{table:SRGN_AGGRE}
\centering
\begin{tabular}{|l|c|ccccc|}
\hline
Dataset & $AGGRE_S(\cdot)$ & F-S & F-D & M-S & M-D & Mean \\
\hline \hline
\multirow{2}*{KinFaceW-I} & Max & \textbf{78.8} &  \textbf{75.4}  & \textbf{80.1}  & 83.8 & \textbf{79.5} \\
 ~ & Avg & 77.6 &  \textbf{75.4}  & 78.0 & \textbf{86.3} & 79.3 \\
\hline
\multirow{2}*{KinFaceW-II} & Max & \textbf{90.8} & 87.0 & \textbf{91.0} & 93.6 & \textbf{90.6} \\
~ & Avg & 90.0 & \textbf{87.8} & 90.4 & \textbf{93.8} & 90.5 \\
\hline
\end{tabular}
\end{table}

\begin{table}[t]
\caption{Mean verification rate (\%) of the H-RGN method using different pooling operations for $AGGRE_H(\cdot)$.}
\label{table:HLRG_AGGRE}
\centering
\begin{tabular}{|l|c|ccccc|}
\hline
Dataset & $AGGRE_H(\cdot)$ & F-S & F-D & M-S & M-D & Mean \\
\hline \hline
\multirow{2}*{KinFaceW-I} & Max &  79.8 &  76.2  & \textbf{81.4} & \textbf{87.8} & 81.3 \\
 ~ & Avg  & \textbf{81.1} &  \textbf{78.4}  & 80.2 & \textbf{87.8} &\textbf{81.9} \\
\hline
\multirow{2}*{KinFaceW-II} & Max & \textbf{91.0} & 85.0 & \textbf{93.0} & 94.6 & 90.9 \\
~ & Avg & \textbf{91.0} & \textbf{85.6} & \textbf{93.0} & \textbf{96.0} & \textbf{91.4} \\
\hline
\end{tabular}
\end{table}

\begin{table}[t]
\caption{The mean verification rate (\%) of the S-RGN and H-RGN methods versus varying $K$.}
\label{table:K}
\centering
\begin{tabular}{|l|c|cccc|}
\hline
\multirow{2}*{Methods} & \multirow{2}*{Dataset} & \multicolumn{4}{|c|}{K}  \\
\cline{3-6}
~ & ~ & 1 & 2 & 3 & 4 \\
\hline \hline
\multirow{2}*{S-RGN} & KinFaceW-I & 74.0 & \textbf{79.5} & 78.7 & 78.0 \\
~ & KinFaceW-II & 86.2 & 90.6 & \textbf{90.8} & 90.1 \\
\hline
\multirow{2}*{H-RGN} & KinFaceW-I & 76.6  & \textbf{81.9} & 81.8 & 81.4 \\
~ & KinFaceW-II & 88.4 & \textbf{91.4} & 91.2 & 91.1 \\
\hline
\end{tabular}
\end{table}

\begin{table}[t]
\caption{The mean verification rate (\%) of the S-RGN and H-RGN methods versus varying $F_1$.}
\label{table:F1}
\centering
\begin{tabular}{|l|c|cccc|}
\hline
\multirow{2}*{Methods} & \multirow{2}*{Dataset} & \multicolumn{4}{|c|}{$F_1$}  \\
\cline{3-6}
~ & ~ & 128 & 256 & 512 & 1024 \\
\hline \hline
\multirow{2}*{S-RGN} & KinFaceW-I & 77.5 & 78.2 & \textbf{79.5} &  78.1  \\
~ & KinFaceW-II & 89.4 & 89.5 & \textbf{90.6} & 89.8 \\
\hline
\multirow{2}*{H-RGN} & KinFaceW-I & 80.9  & \textbf{82.0} & 81.9 & 81.6 \\
~ & KinFaceW-II & 89.9 & 90.4 & \textbf{91.4} & 91.3 \\
\hline
\end{tabular}
\end{table}

\begin{table}[t]
\caption{The mean verification rate (\%) of the S-RGN and H-RGN methods versus varying $F_2$.}
\label{table:F2}
\centering
\begin{tabular}{|l|c|cccc|}
\hline
\multirow{2}*{Methods} & \multirow{2}*{Dataset} & \multicolumn{4}{|c|}{$F_2$}  \\
\cline{3-6}
~ & ~ & 1 & 2 & 4 & 8 \\
\hline \hline
\multirow{2}*{S-RGN} & KinFaceW-I & 74.5 & 78.1 & \textbf{79.5} & 79.4 \\
~ & KinFaceW-II & 88.7 & 89.4 & \textbf{90.6} & 90.1 \\
\hline
\multirow{2}*{H-RGN} & KinFaceW-I & 80.8  & 81.7 & 81.9 & \textbf{82.3} \\
~ & KinFaceW-II & 90.0 & 90.2 & \textbf{91.4} & 91.2 \\
\hline
\end{tabular}
\vspace{-0.5cm}
\end{table}

\emph{Parameter Analysis:} We first tested the mean verification rate of the proposed S-RGN and H-RGN methods on the KinFaceW-I database and  KinFaceW-II database with different parameters and design choices and then applied these parameters and design choices for all following experiments.

Both the S-RGN and H-RGN introduced latent reasoning nodes, which need to be initialized before the message passing stage. Different pooling operations are considered for $INIT_S(\cdot)$ and we further propose a self-attention mechanism for $INIT_H(\cdot)$ to extract more discriminative initial features. Tables \ref{table:SRGN_INIT} and \ref{table:HLRG_INIT} show the results with different implements of $INIT_S(\cdot)$ and $INIT_H(\cdot)$, respectively. We observe that average pooling always gives better performances than mean pooling for $INIT_S(\cdot)$ and $INIT_H(\cdot)$ and the proposed self-attention scheme is superior to these two pooling operations, which illustrates that the self-attention mechanism learns a more meaningful initialization. In addition, the S-RGN method is very sensitive to the choice of node initialization function, while the H-RGN method is more robust. The main reason is that the H-RGN introduces multiple layers of latent nodes to aggregate the information of the underlying nodes hierarchically. On the other hand, the S-RGN only introduces one hidden node as the information communication bridge between all the other nodes, so the central node has a significant impact on the performance.

To update the node features, the messages from other nodes are first aggregated with a mapping function in our S-RGN and H-RGN methods. We consider different pooling operations to implement the $AGGRE_S(\cdot)$ and $AGGRE_H(\cdot)$ functions. The results of the verification rate on the KinFaceW-I and KinFaceW-II datasets are tabulated in Table \ref{table:SRGN_AGGRE} and \ref{table:HLRG_AGGRE}. We see that the performance difference between different pooling methods for aggregate functions is small. For S-RGN, max-pooling gives slightly better performance, while max-pooling achieves slightly better results for the H-RGN method. One possible explanation for why max pooling is more suitable for the S-RGN method is that the star structure requires the central node to extract information through one aggregation operation. Max pooling can help the S-RGN method to select more important information, while mean pooling treats all messages equally. On the contrary, the hierarchical structure of the H-RGN method can effectively extract important information, and mean pooling can better extract detailed and comprehensive information. Therefore, we adopted max pooling for $AGGRE_S(\cdot)$ and mean pooling for $AGGRE_H(\cdot)$ in the following experiments.

We analyse the effect of $K$ in our S-RGN and H-RGN methods. To vary $K$, we set $F_K = 4, F_1 = ... = F_{K-1} = 512$. Table \ref{table:K} lists the mean verification rate of the S-RGN and H-RGN methods versus different values of $K$. The results show that the $K$ should be set as $2$ to obtain the best mean verification rate in most cases. Having set $K$ as 2, we consider the setting of feature dimensions of the two-layer S-RGN networks and H-RGN networks. We first fix the feature dimension of the second layer of S-RGN and H-RGN networks to 4($F_2 = 4$), and then observe the experimental results on the KinFaceW-I and KinFaceW-II databases versus different values of the first layer feature dimension $F_1$. We notice that both S-RGN and H-RGN methods achieve satisfactory performance when $F_1$ is 512 as shown in Table \ref{table:F1}. Subsequently, we examine the mean verification rate versus varying $F_2$ by fixing $F_1 = 512$. The results are shown in Table \ref{table:F2}, and the parameter $F_2$ is selected as $4$, which reaches the best performance in most cases.

\begin{table}[t]
\caption{The verification rate (\%) of the H-RGN method with different graph configurations. }
\label{table:HLRG_nodes}
\centering
\begin{tabular}{|l|c|ccccc|}
\hline
\multirow{2}*{$L$} &  \# of nodes  & \multirow{2}*{F-S} & \multirow{2}*{F-D} & \multirow{2}*{M-S} & \multirow{2}*{M-D} & \multirow{2}*{Mean} \\
 ~ &  per layer & ~ & ~ &  ~ & ~ & ~ \\
\hline \hline
\multirow{4}*{1} & (1,) & 80.1  &  76.9  & 79.3 & 88.6 & 81.2 \\
 ~ & (16,) & 80.1 & 78.8 & 78.9 & 87.4 & 81.3 \\
 ~ & \textbf{(32,)} & 79.8 &  76.9  & 81.0 & 88.2 & \textbf{81.5} \\
 ~ & (64,) & 80.4 &  74.7  & 79.2 & 88.6 & 80.7 \\
\hline
\multirow{7}*{2} & (128,1) & 80.2  &  77.3  & 80.1 & 88.2 & 81.5 \\
 ~ & \textbf{(128,16)} & 80.2 & 78.4 & 80.6 & 89.8 & \textbf{82.3} \\
 ~ & (128,32) & 82.1 &  77.7  & 79.3 & 88.6 & 81.9 \\
 ~ & (128,64) & 81.1 &  76.9  & 78.9 & 88.6 & 81.4 \\
 \cline{2-7}
 ~ & (256,16) & 80.5 & 79.1 & 80.1 & 88.2 & \textbf{82.0} \\
 ~ & (256,32) & 81.8 &  77.3  & 79.7 & 88.9 & 81.9 \\
 ~ & (256,64) & 81.8 &  76.5  & 80.2 & 88.2 & 81.7 \\
\hline
\multirow{5}*{3} & (128,32,1) & 83.0  &  76.2  & 80.6 & 89.0 & 82.2 \\
 ~ & \textbf{(128,32,8)} & 81.7 &  78.8  & 81.4 & 88.6 & \textbf{82.6} \\
 ~ & (128,64,1) & 80.4 & 77.6 & 82.7 & 87.4 & 82.0 \\
 ~ & (128,64,16) & 80.8 &  78.4  & 80.2 & 89.0 & 82.1 \\
 ~ & (256,64,1) & 81.1 &  78.4  & 80.2 & 87.8 & 81.9 \\
 \hline
 \multirow{2}*{4} & \textbf{(128,32,8,2)} & 80.5  &  77.7  & 81.0 & 88.2 & \textbf{81.9} \\
 ~ & (256,64,16,4) & 80.8 & 76.2 & 79.7 & 88.2 & 81.2 \\
\hline
\end{tabular}
\vspace{-0.5cm}
\end{table}

Lastly, we study how H-RGN uses hierarchical reasoning graphs to improve reasoning ability. More specifically, we conducted experiments on the KinFaceW-I database with different graph configurations and the results are shown in Table \ref{table:HLRG_nodes}.

We first analyse the effect of the number  of latent layers $L$ and find that appropriately increasing the number of latent layers is beneficial to performance. We observe that adding a latent layer with 128 or 256 nodes to the graph with only one latent layer can improve the performance. For example, the graph configurations of (128,16) and (256,16) achieve the mean verification rate of 82.3\% and 82.0\%, respectively, and they both give better results than the graph configuration of (16,). Moreover, as shown in the table, introducing another latent layer to the graphs with two latent layers can further boost performance, which illustrates that more reasoning capacity can be exploited through the hierarchical latent graph. We also notice that a deeper graph structure with four latent layers may lead to over-fitting and the best performance is obtained with three latent reasoning layers.

We then analyse the impact of the number of nodes in each latent layer. We see that when $L=1$, the graph configuration (32,) achieves the best result; when $L=2$, the graph structure (128,16) is the most excellent structure; when $L=3$, the nodes setting (128,32,8) is superior to the others; when $L=4$, the graph configuration (128,32,8,2) gives better performance than the structure (256,64,16,4). It should be pointed out that the hierarchical graph also contains the $0^{th}$ layer with $D=512$ visual comparison nodes, which results in the actual number of layers of our graph being $L+1$. We observe that these superior structures have a moderate node reduction rate between layers. In particular, for the superior graph configurations with $L > 1$, the number of nodes in a layer decreases by 4 $\sim$ 8 times. The advantages of these structures may be due to their better balance between the comprehensiveness and locality in the information aggregation process.  We take these superior structures as the default structures for the corresponding latent layer number in the subsequent experiments.

\begin{figure*}[t]
  \centering
  \subfigure[]{
    \label{fig:roc:kfw1}
    \includegraphics[width=0.31\linewidth]{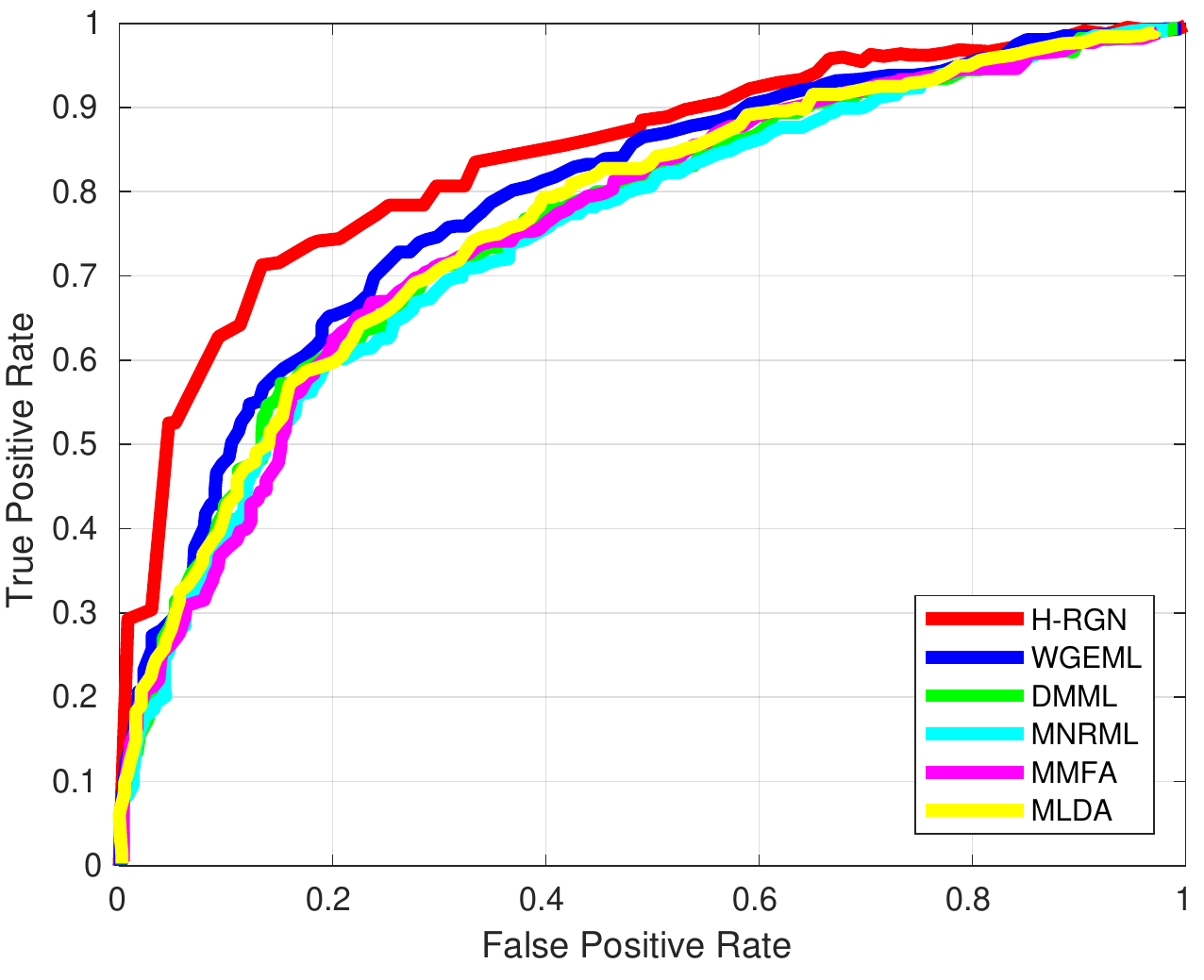}
  }
  \subfigure[]{
    \label{fig:roc:kfw2}
    \includegraphics[width=0.31\linewidth]{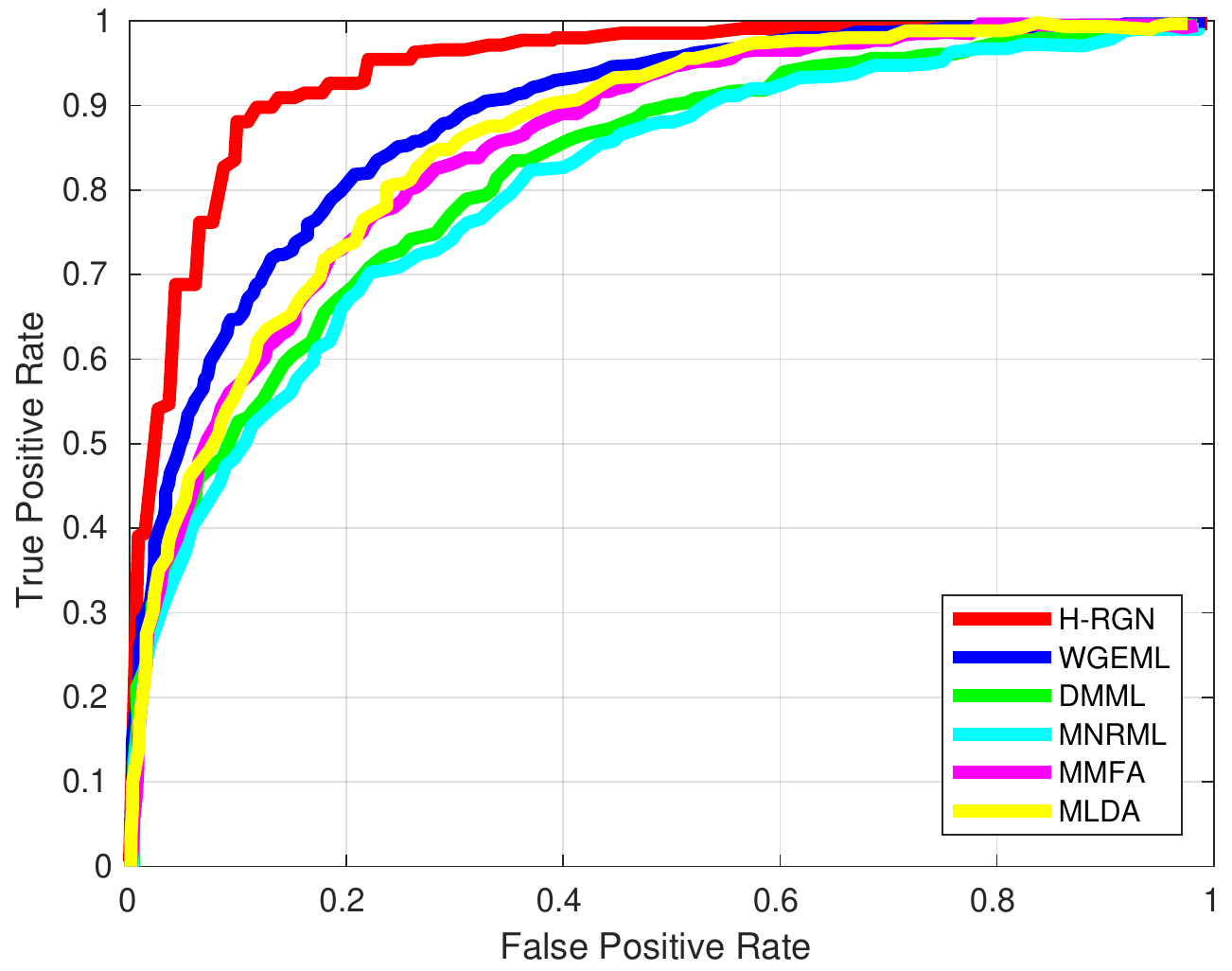}
  }
  \subfigure[]{
    \label{fig:roc:tskin}
    \includegraphics[width=0.31\linewidth]{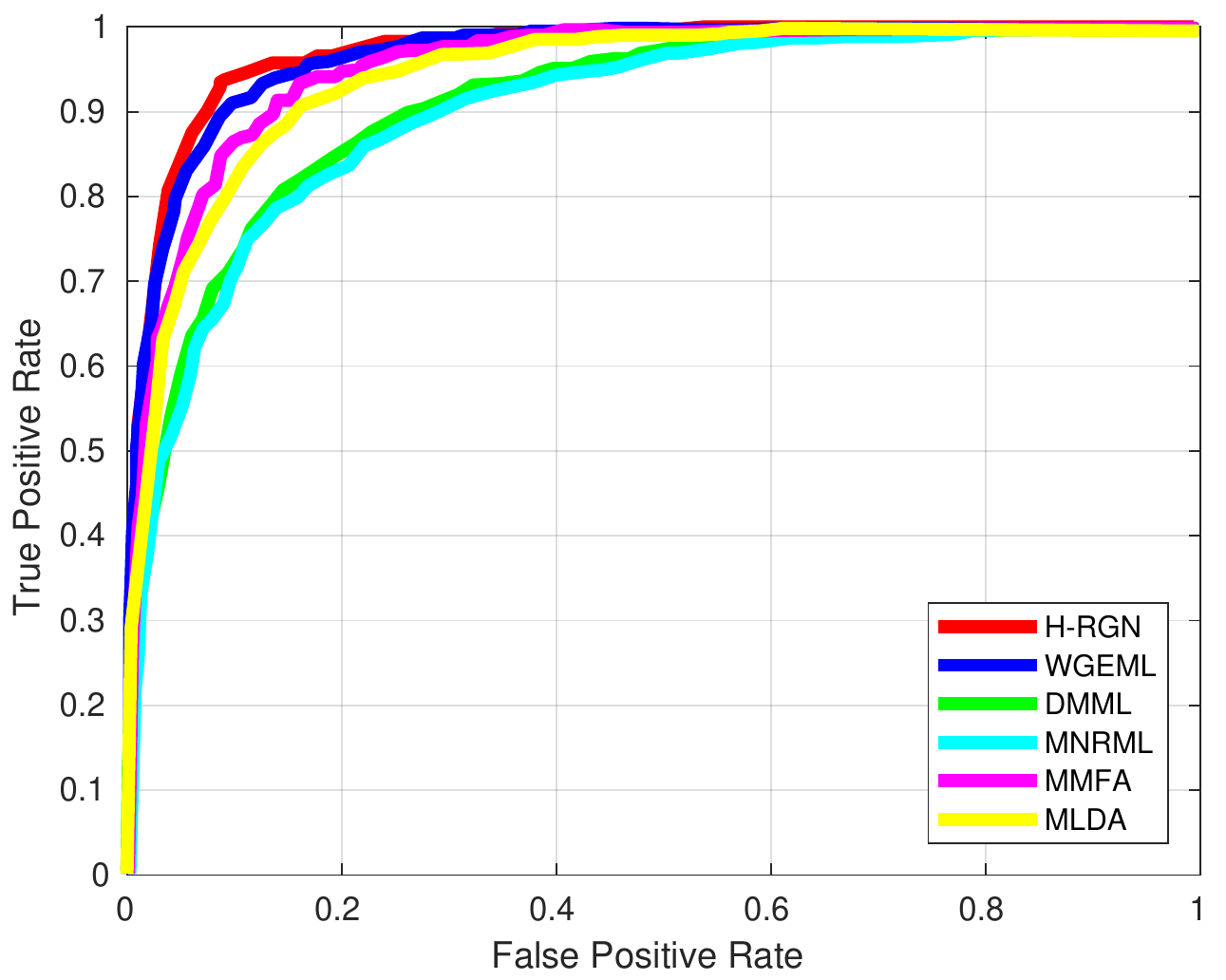}
  }
  \caption{The ROC curves of different methods obtained on the (a) KinFaceW-I, (b) KinFaceW-II, and (c) TSKinFace databases, respectively.}
  \label{fig:roc}
  \vspace{-0.5cm}
\end{figure*}

\begin{table*}[t]
\caption{Performance comparisons (\%) with other methods on the KinFaceW-I and KinFaceW-II datasets.}
\label{table:KFWIandII}
\renewcommand\tabcolsep{9pt}
\centering
\begin{tabular}{|l|ccccc|ccccc|}
\hline
\multirow{2}*{Method} & \multicolumn{5}{|c|}{KinFaceW-I} & \multicolumn{5}{c|}{KinFaceW-II}\\
\cline{2-11}
~ & F-S & F-D & M-S & M-D & Mean & F-S & F-D & M-S & M-D & Mean\\
\hline \hline

MNRML~\cite{lu2013neighborhood} & 72.5 & 66.5 & 66.2 & 72.0 & 69.9 & 76.9 & 74.3 & 77.4 & 77.6 & 76.5 \\
MPDFL~\cite{yan2014prototype} & 73.5 & 67.5 & 66.1 & 73.1 & 70.1 & 77.3 & 74.7 & 77.8 & 78.0 & 77.0 \\
DMML~\cite{yan2014discriminative} & 74.5 & 69.5 & 69.5 & 75.5 &  72.3 & 78.5 & 76.5 & 78.5 & 79.5 & 78.3 \\
DDML~\cite{lu2017discriminative} & 73.3 & 66.1 & 73.5 & 72.6 & 71.4 & 77.0 & 70.4 & 74.4 & 76.8 & 74.7 \\
L$^2$M$^3$L~\cite{hu2017local} & - & - & - & - & - & 82.4 & 78.2 & 78.8 & 80.4 & 80.0 \\
MvDML~\cite{hu2018sharable} & - & - & - & - & - & 80.4 & 79.8 & 78.8 & 81.8 &  80.2 \\
D-CBFD~\cite{yan2019learning} & 79.0 & 74.2 & 75.4 & 77.3 & 78.5 & 81.0 & 76.2 & 77.4 & 79.3 & 78.5 \\
WGEML~\cite{liang2018weighted} & 78.5 & 73.9 & 80.6 & 81.9 & 78.7 & 88.6 & 77.4 & 83.4 & 81.6 & 82.8 \\
MHDL~\cite{mahpod2018kinship} & 77.0 & 76.1 & 80.2 & 85.9 & 79.8 & 88.4 & 84.0 & 86.4 & 89.2 & 87.0 \\
\hline
CNN-Basic~\cite{BMVC2015_148} & 75.7 & 70.8 & 73.4 & 79.4 & 74.8 & 84.9 & 79.6 & 88.3 & 88.5 & 85.3 \\
CNN-Point~\cite{BMVC2015_148} & 76.1 & 71.8 & 78.0 & 84.1 & 77.5 & 89.4 & 81.9 & 89.9 & 92.4 & 88.4 \\
CFT*~\cite{duan2017face} & 78.8 & 71.7 & 77.2 & 81.9 & 77.4 & 77.4 & 76.6 & 79.0 & 83.8 & 79.2 \\
NESN-KVN~\cite{wang2020discriminative} & 77.0 & 76.5 & 75.8 & 85.2 & 78.6 & 88.7 & 86.7 & 89.1 & 91.6 & 89.0 \\
AdvKin~\cite{zhang2020advkin} & 75.7 & 78.3 & 77.6 & 83.1 & 78.7 & 88.4 & 85.8 & 88.0 & 89.8 & 88.0 \\
\hline \hline
Baseline-MLP & 70.9 & 68.3 & 75.0 & 80.4 & 73.7 & 77.6 & 78.6 & 78.2 & 79.8 & 78.6 \\
Baseline-Cos & 75.0 & 75.0 & 78.4 & 82.6 & 77.8 & 82.8 & 80.6 & 83.2 & 86.4 & 83.0 \\
\hline \hline
S-RGN & 78.8 &  75.4  & 80.1  & 83.8 & 79.5 & \textbf{90.8} & 87.0 & 91.0 & 93.6 & 90.6 \\
\hline
H-RGN (L=1) & 79.8 &  76.9  & 81.0 & 88.2 & 81.5  & 90.2 & 86.6 & 92.4 & 95.2 & 91.1 \\
H-RGN (L=2) & 80.2 & 78.4 & 80.6 & \textbf{89.8} & 82.3 & 90.4 & \textbf{87.2} & \textbf{93.6} & 95.8 & \textbf{91.8} \\
H-RGN (L=3) & \textbf{81.7} &  \textbf{78.8}  & \textbf{81.4} & 88.6 & \textbf{82.6} & 90.6 & 86.8 & 93.0 & \textbf{96.0} & 91.6 \\
\hline
\end{tabular}
\vspace{-0.5cm}
\end{table*}

\emph{Comparison with the State-of-the-arts:} Table \ref{table:KFWIandII} presents the comparison performance with different approaches on the KinFaceW-I and KinFaceW-II datasets. We observe that the S-RGN achieves a mean verification accuracy of 79.5\% on the KinFaceW-I dataset and that of 90.6\% on the KinFaceW-II dataset, which is competitive with the state-of-the-art methods. The H-RGN with two latent layers further improves the performance to 82.3\% and 91.8\% on KinFaceW-I and KinFaceW-II datasets respectively, which outperforms state-of-the-art methods. Most existing state-of-the-art methods can be grouped into metric-learning based methods and deep learning-based methods. Some early metric learning-based approaches, such as MNRML~\cite{lu2013neighborhood} and DMML~\cite{yan2014discriminative} learn a metric with hand-crafted features, which leads to unsatisfactory performance. The methods of WGEML~\cite{liang2018weighted} and MHDL~\cite{mahpod2018kinship} achieve state-of-the-art results with deeply learned features, which illustrates the superiority of deep learning. Compared with MHDL, The H-RGN (L=2) improves the mean accuracy by 2.5\% and 4.8\% on KinFaceW-I and KinFaceW-II datasets, respectively, which shows the superior relational reasoning ability of our method. Zhang \emph{et al.}~\cite{BMVC2015_148} proposed the CNN-Basic and CNN-Point, which was the first attempt to leverage learn deep neural networks for kinship verification. Our methods, which are also deep learning-based methods, significantly outperforms the CNN-Point by a large margin on both datasets. Note that CNN-Point has 10 CNN backbones while our methods only utilize one CNN backbone, which further demonstrates the effectiveness of the proposed methods. We visualize the receiver operating characteristic (ROC) curves of different methods in Fig. \ref{fig:roc} to make an intuitive comparison, where Fig. \ref{fig:roc:kfw1} and \ref{fig:roc:kfw2} plot the ROC curves of the results on the KinFaceW-I and KinFaceW-II data sets, respectively. We see that our method yields the best performance.

\emph{Ablation Study:} To validate the effectiveness of our proposed methods, we compare them with two baseline methods. These baseline methods employ the same CNN backbone to extract image features and the only difference from our methods is the mapping function $f(\cdot)$. We observe that the S-RGN achieves better results on both datasets than the two baseline methods. Especially on the KinFaceW-II dataset, the S-RGN method improves the baseline performance by $7.6\% \sim 12.0\%$, which demonstrates that the commonly used MLP and cosine similarity schemes cannot effectively discover the hidden similarity between two identities, while our method can better exploit the genetic relation with a reasoning graph. Moreover, we see that the H-RGN further improves the performance of S-RGN, which shows the superiority of our hierarchical reasoning graph over the star-shaped graph.

\begin{table*}[t]
\caption{Performance comparisons (\%) with other methods on the TSKinFace Dataset.}
\renewcommand\tabcolsep{9pt}
\label{table:TSKinFace}
\centering
\begin{tabular}{|l|ccccc|ccc|c|}
\hline
\multirow{2}*{Method} &  \multicolumn{5}{|c|}{Bi-subject}  &  \multicolumn{3}{|c|}{Tri-subject} & Mean of all \\
\cline{2-9}
~ & F-S & F-D & M-S & M-D & Mean & FM-S & FM-D & Mean & six relations \\
\hline \hline
MNRML~\cite{lu2013neighborhood} & 83.2 & 81.4 & 83.2 & 82.1 & 82.5 & 87.1 & 85.7 & 86.4 & 83.8 \\
DMML~\cite{yan2014discriminative} & 83.8 & 81.8 & 84.1 & 82.4 & 83.0 & 87.9 & 86.1 & 87.0 & 84.3 \\
RSBM-block-FS~\cite{qin2015tri} & 83.0 & 80.5 & 82.8 & 81.1 & 81.9 & 86.4 & 84.4 & 85.4 & 83.0 \\
GMP~\cite{zhang2015group} & 88.5 & 87.0 & 87.9 & 87.8 & 87.8 & 90.6 & 89.0 & 89.8 & 88.5 \\
DDML~\cite{lu2017discriminative} & 75.7 & 74.6 & 76.4 & 78.1 & 76.2 & 78.8 & 80.1 & 79.5 & 77.3 \\
LC-FS~\cite{zhang2016genetics} & - & - & - & - & - & 91.1 & 88.3 & 89.7 & - \\
MKSM~\cite{zhao2018learning} & 84.8 & 83.2 & 85.2 & 84.9 & 84.5 & - & - & - & - \\
MSIDA~\cite{bessaoudi2019multilinear} & - & - & - & - & 85.2 & - & - & - & - \\
WGEML~\cite{liang2018weighted} & 90.3 & 89.8 & 91.4 & 90.4 & 90.5 & 93.5 & 93.0 & 93.3 & 91.4 \\
\hline \hline
Baseline-MLP & 76.5 & 74.5 & 79.5 & 79.2 & 77.4 & 80.2 &  78.3 & 79.3 & 78.0 \\
Baseline-Cos & 82.1 & 80.6 & 82.3 & 81.3 & 81.6 & 85.2 &  83.7 & 84.5 & 82.5 \\
\hline \hline
S-RGN & 91.3 &  87.9 & 91.8  & 91.6& 90.7 & 91.8 &  \textbf{94.5} & 93.2 & 91.5  \\
\hline
H-RGN (L=1) & 91.8 &  90.7  & 91.8 & 92.5 & 91.7 & 94.2 &  93.3 & 93.8 & 92.4   \\
H-RGN (L=2) & \textbf{92.0} & \textbf{91.5} & \textbf{93.0} & \textbf{92.8} & \textbf{92.3} & \textbf{94.5} &  93.2 & 93.9 & \textbf{92.8}  \\
H-RGN (L=3) &  \textbf{92.0}  & 90.6 & 92.8 & 92.0 & 91.9  & 94.2 &  93.7 & \textbf{94.0} & 92.6  \\
\hline
\end{tabular}
\vspace{-0.5cm}
\end{table*}

\subsection{Experimental Results on TSKinFace}
Although our methods are specially designed for the kinship involving only two subjects (one-versus-one) such as father-daughter or mother-son, we have conducted experiments on the TSKinFace dataset to verify that they can also be applied to tri-subject (one-versus-two) kinship learning with some minor modifications. The  tri-subject kinship verification aims to investigate the relationships among multiple visual entities and answer the question of whether a child in an image belongs to given parents. To extend our methods to tri-subject kinship verification, we first use the same deep CNN $g(\cdot)$ to extract features from three facial images, and then let each visual comparison node model the comparison of three individuals in one feature dimension. Subsequently, the initial features of the $D$ visual comparison node are set as the concatenation of the values of the three extracted deep features in the same dimension. Having obtained the initial values of $D$ visual comparison nodes, the following operations can directly follow the framework of bi-subject kinship verification. Table \ref{table:TSKinFace} shows the comparison with different methods on the TSKinFace dataset. To make a more comprehensive comparison, we also show the results of one-versus-one kinship verification.

\emph{Comparison with the State-of-the-arts:} We first analyse the results of bi-subject kinship verification. We observe that the S-RGN method achieves a mean verification accuracy of 90.7\%.  The H-RGN method further improves the performance to 92.3\% with two latent layers, outperforming the state-of-the-art~\cite{liang2018weighted} by 1.8\%. In addition, we see that the performance of tri-subject kinship recognition is generally higher than that of bi-subject kinship verification, which is reasonable considering that tri-subject kinship reasoning provides more information about parents. Concretely, for tri-subject kinship verification, the S-RGN attains 93.2\% mean accuracy, which is comparable with existing methods~\cite{liang2018weighted,zhang2016genetics}. The H-RGN method with two latent layers reaches the performance of 93.9\% mean accuracy, which is superior to state-of-the-art methods~\cite{liang2018weighted,zhang2016genetics}. These consistently superior results verify that our methods are not only suitable for one-versus-one kin recognition, but also can be successfully applied to one-versus-two kinship learning. Fig. \ref{fig:roc:tskin} presents the ROC results on the TSKinFace database and we observe that our method achieves very competitive results.

\emph{Ablation Study:} Compared with Baseline-MLP and Baseline-Cos, the S-RGN method attains 13.3\% and 9.1\%  improvements for bi-subject kinship recognition, and 13.9\% and 8.7\% improvements for tri-subject kinship learning, which consistently demonstrates the effectiveness of the graph reasoning module. We see that extending the star-shaped graph in the S-RGN method to a hierarchical graph with one latent layer can improve the performance of bi-subject and tri-subject kinship recognition by 1.0\% and 0.6\%, respectively.  It demonstrates that more reasoning capacity can be exploited by a hierarchical reasoning graph. Finally, the H-RGN with two latent reasoning layers gives the highest overall performance of 92.8\% mean accuracy of all six kinship relations.

\begin{figure}[t]
  \centering
  \includegraphics[width=0.85\linewidth]{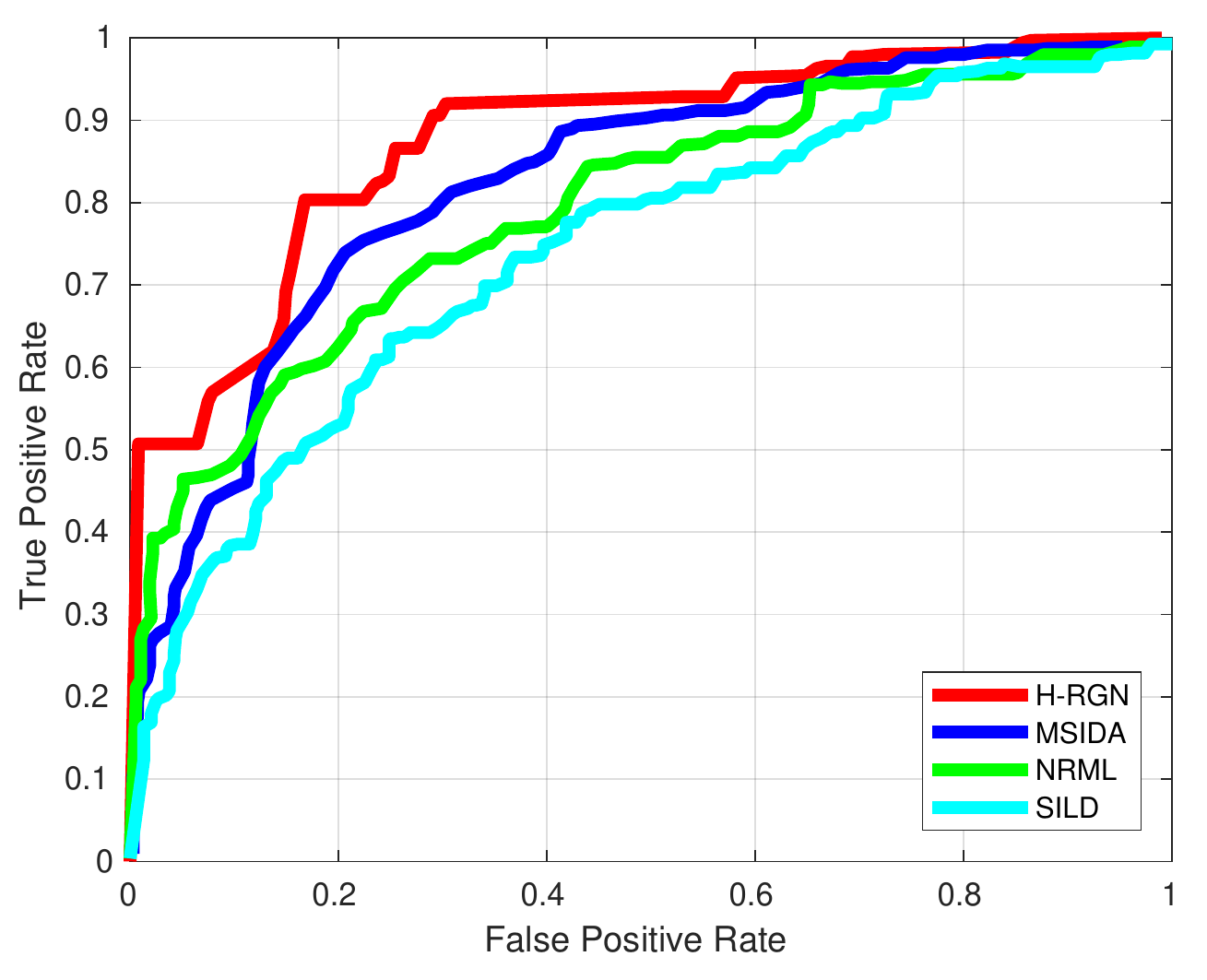}
  \caption{The ROC curves of different methods on the Cornell KinFace Dataset.}
  \label{fig:roc:cornell}
  \vspace{-0.5cm}
\end{figure}

\subsection{Experimental Results on the Cornell KinFace Dataset}
We further conducted experiments on the Cornell KinFace dataset. Table \ref{table:Cornell} presents the performance comparisons with different methods.

\emph{Comparison with the State-of-the-arts:} We see that S-RGN and H-RGN (L=3) achieve the mean verification rate of 87.4\% and 89.6\%, respectively. Both of them outperform the state-of-the-art approaches, which illustrates the effectiveness of our proposed methods. Compared with the recent deep learning-based methods CFT*~\cite{duan2017face} and AdvKin~\cite{zhang2020advkin}, our H-RGN method with three latent layers obtains 11.3\% and 8.2\% improvements. Moreover, our H-RGN outperforms the state-of-the-art~\cite{bessaoudi2019multilinear} by 2.7\%. We show the ROC curves of different approaches in Fig. \ref{fig:roc:cornell}. It is clear that the ROC curves of our proposed H-RGN are higher than those of other methods.

\begin{table}[t]
\caption{Performance comparisons (\%) with other methods on the Cornell KinFace Dataset.}
\renewcommand\tabcolsep{8pt}
\label{table:Cornell}
\centering
\begin{tabular}{|l|ccccc|}
\hline
Method & F-S & F-D & M-S & M-D & Mean \\
\hline \hline
MNRML~\cite{lu2013neighborhood} & 74.5 & 68.8 & 77.2 & 65.8 & 71.6 \\
DMML~\cite{yan2014discriminative} & 76.0 & 70.5 & 77.5 & 71.0 & 73.8 \\
MPDFL~\cite{yan2014prototype} & 74.8 & 69.1 & 77.5 & 66.1 & 71.9 \\
SILD~\cite{kan2011side} & - & - & - & - & 71.4 \\
KML~\cite{zhou2019learning} & 78.9 & 82.6 & 78.3 & 85.7 & 81.4 \\
MKSM~\cite{zhao2018learning} & 80.5 & 80.6 & 79.5 & 86.2 & 81.7 \\
MSIDA~\cite{bessaoudi2019multilinear} & - & - & - & - & 86.9 \\
\hline
CFT*~\cite{duan2017face} & - & - & - & - & 78.3 \\
AdvKin~\cite{zhang2020advkin} & - & - &  - &  - & 81.4 \\
\hline \hline
Baseline-MLP & 78.9 & 80.0 & 84.1 & 87.5 & 82.6 \\
Baseline-Cos & 82.0 & 77.1 & 93.3 & 91.7 & 86.0 \\
\hline \hline
S-RGN & 75.0 &  82.9 & 95.0  & \textbf{96.7} & 87.4  \\
\hline
H-RGN (L=1) & 82.1 &  \textbf{85.7}  & 94.2 & 91.7 & 88.4  \\
H-RGN (L=2) & 83.6 & 82.9 & 94.2 & 95.0 & 88.9  \\
H-RGN (L=3) &  \textbf{84.4}  & \textbf{85.7} & \textbf{97.5} & 90.8 & \textbf{89.6}  \\
\hline
\end{tabular}
\vspace{-0.3cm}
\end{table}

\emph{Ablation Study:} When we apply the graph reasoning module in the face matching stage with S-RGN, the mean verification rate increase 4.8\% and 1.4\% over Baseline-MLP and Baseline-Cos respectively, which illustrates its effectiveness. The reasoning ability of S-RGN is limited by the star structure, which can be alleviated by the hierarchical structure. We observe that one-layer H-RGN attains 1.0\% improvements and the H-RGN method with three latent layers obtains the mean verification rate of 89.6\%, outperforming the S-RGN by 2.2\%.

\begin{table}[t]
\caption{Performance and Computational Complexity with Different Latent Layers.}
\renewcommand\tabcolsep{4pt}
\label{table:mac}
\centering
\begin{tabular}{|l|ccccc|}
\hline
layer number $L$ & 1 & 2 & 3 & 4 & 5 \\
\hline \hline
Verification rate (\%)  & 81.5 & 82.3 & \textbf{82.6} & 81.9 & 81.8 \\
MACs (M) & \textbf{584.12} & 710.53 & 740.02 & 745.27 & 926.52 \\
\hline
\end{tabular}
\vspace{-0.5cm}
\end{table}

\subsection{Computational Complexity with Different Latent Layers}
To show the computational complexity with different latent layers, we conducted experiments on the KinFaceW-I database with different numbers of latent layers. The results are presented in Table \ref{table:mac}. We report the multiply-accumulate (MAC) operations to demonstrate the computational complexity. Besides, we also attach the mean verification rate with different layers. We observe that increasing the number of layers appropriately improves the performance. When the number of layers reaches four, our reasoning graph networks may lead to over-fitting. We also see that the MACs increase with the number of layers, which is reasonable since more layers introduce more computation. The trade-off between performance and computational cost can be determined according to one's own needs.

\begin{table}[t]
\caption{Performance comparisons (\%) with different face datasets pre-trained backbone networks.}
\renewcommand\tabcolsep{7pt}
\label{table:faces}
\centering
\begin{tabular}{|l|ccccc|}
\hline
Face dataset & F-S & F-D & M-S & M-D & Mean \\
\hline \hline
None & 81.7 & 78.8 & 81.4 & \textbf{88.6} & 82.6 \\
VGG-Face2 & \textbf{84.6} & 80.2 & \textbf{82.7} & 88.2 & \textbf{83.9} \\
CASIA-WebFace & 84.0 & \textbf{81.7} & \textbf{82.7} & 87.0 & \textbf{83.9} \\
\hline
\end{tabular}
\vspace{-0.5cm}
\end{table}

\subsection{Pre-training with Additional Face Datasets}
This paper mainly focuses on the face matching stage and considers how to compare and fuse two image features. Therefore, we only use the ImageNet pre-trained feature extractor network  to show the effectiveness of our proposed method. Even without the additional face datasets, our method achieves state-of-the-art results, which demonstrates the superiority of our method. To further advance the performance, we can use additional large-scale face datasets (such as VGG-Face2~\cite{cao2018vggface2}, CASIA-WebFace~\cite{yi2014learning}) to pre-train the backbone network as many other methods~\cite{lu2017discriminative,zhang2020advkin} did. we conducted experiments with a three-layer hierarchical reasoning graph network on the KinFaceW-I database. Different large-scale face datasets are used to pre-train the backbone network and the results are shown in Table~\ref{table:faces}. We observe that the mean verification rate is further improved to 83.9\%, which shows the benefits of outside data training.

\section{Conclusions}

In this paper, we have presented a star-shaped reasoning graph network for kinship verification. Our method can effectively reason with two extract image features while most existing methods fail to explicitly model the reasoning process.  Moreover, we have extended our  star-shaped reasoning graph networks to hierarchical reasoning graph networks, which demonstrate a more powerful and flexible reasoning capacity. Extensive experimental results show that our methods achieve superior performance compared with the state-of-the-art methods. Besides, we have verified that our methods are also suitable for tri-subject kinship verification. How to apply our methods to other visual applications such as face verification is a proposing direction of our future work.

{
\bibliographystyle{splncs}
\bibliography{IEEEabrv,rgn}
}

\begin{IEEEbiography}[{\includegraphics[width=1in,height=1.25in,clip,keepaspectratio]{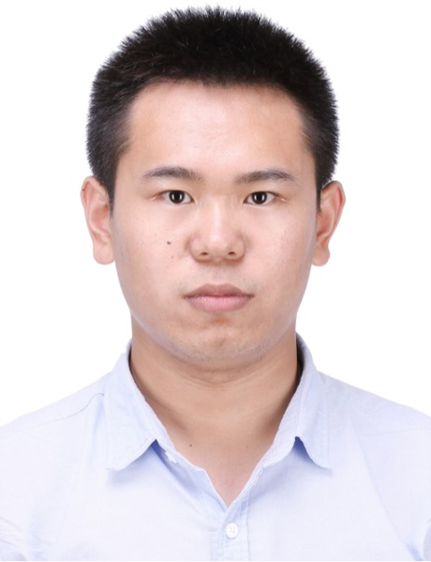}}]{Wanhua Li}
received the B.S. degree from the School of Data and Computer Science, Sun Yat-Sen University, Guangzhou, China, in 2017.
He is currently a Ph.D Candidate with the Department of Automation, Tsinghua University, China. His research interests include facial attribute analysis and graph neural networks. He serves as a regular reviewer member for a number of journals and conferences, \emph{e.g.}, IEEE Transactions on Image Processing, IEEE Transactions on Circuits and Systems for Video Technology, Neural Networks, Neurocomputing, International Conference on Computer Vision and so on.
\end{IEEEbiography}

\begin{IEEEbiography}[{\includegraphics[width=1in,height=1.25in,clip,keepaspectratio]{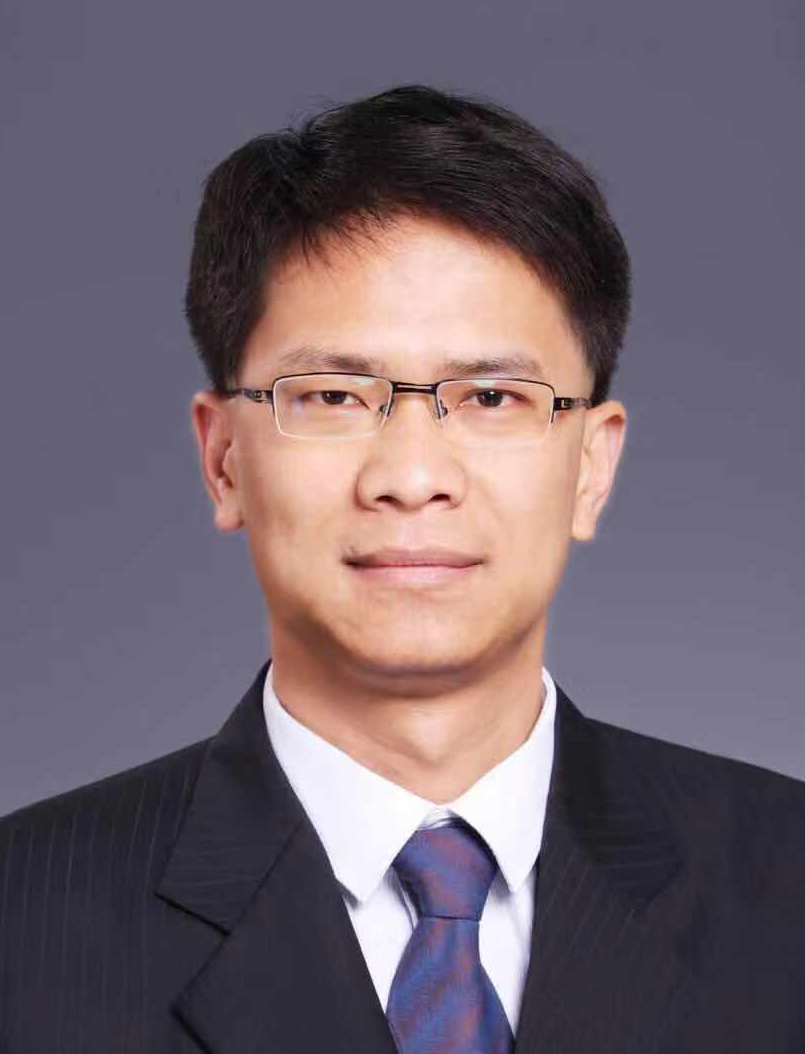}}]{Jiwen Lu}
(M'11-SM'15) received the B.Eng. degree in mechanical engineering and the M.Eng. degree in electrical engineering from the Xi'an University of Technology, Xi'an, China, in 2003 and 2006, respectively, and the Ph.D. degree in electrical engineering from Nanyang Technological University, Singapore, in 2012. He is currently an Associate Professor with the Department of Automation, Tsinghua University, Beijing, China. His current research interests include computer vision, pattern recognition, and intelligent robotics, where he has authored/co-authored over 270 scientific papers in these areas. He serves the Co-Editor-of-Chief of the Pattern Recognition Letters, an Associate Editor of the IEEE Transactions on Image Processing, the IEEE Transactions on Circuits and Systems for Video Technology, the IEEE Transactions on Biometrics, Behavior, and Identity Science, and the Pattern Recognition journal. He also serves as the General Co-Chair of IEEE ICME'2022, and the Program Co-Chair of IEEE FG'2023, IEEE VCIP'2022, IEEE AVSS'2021 and IEEE ICME'2020. He is an IAPR Fellow.
\end{IEEEbiography}

\begin{IEEEbiography}[{\includegraphics[width=1in,height=1.25in,clip,keepaspectratio]{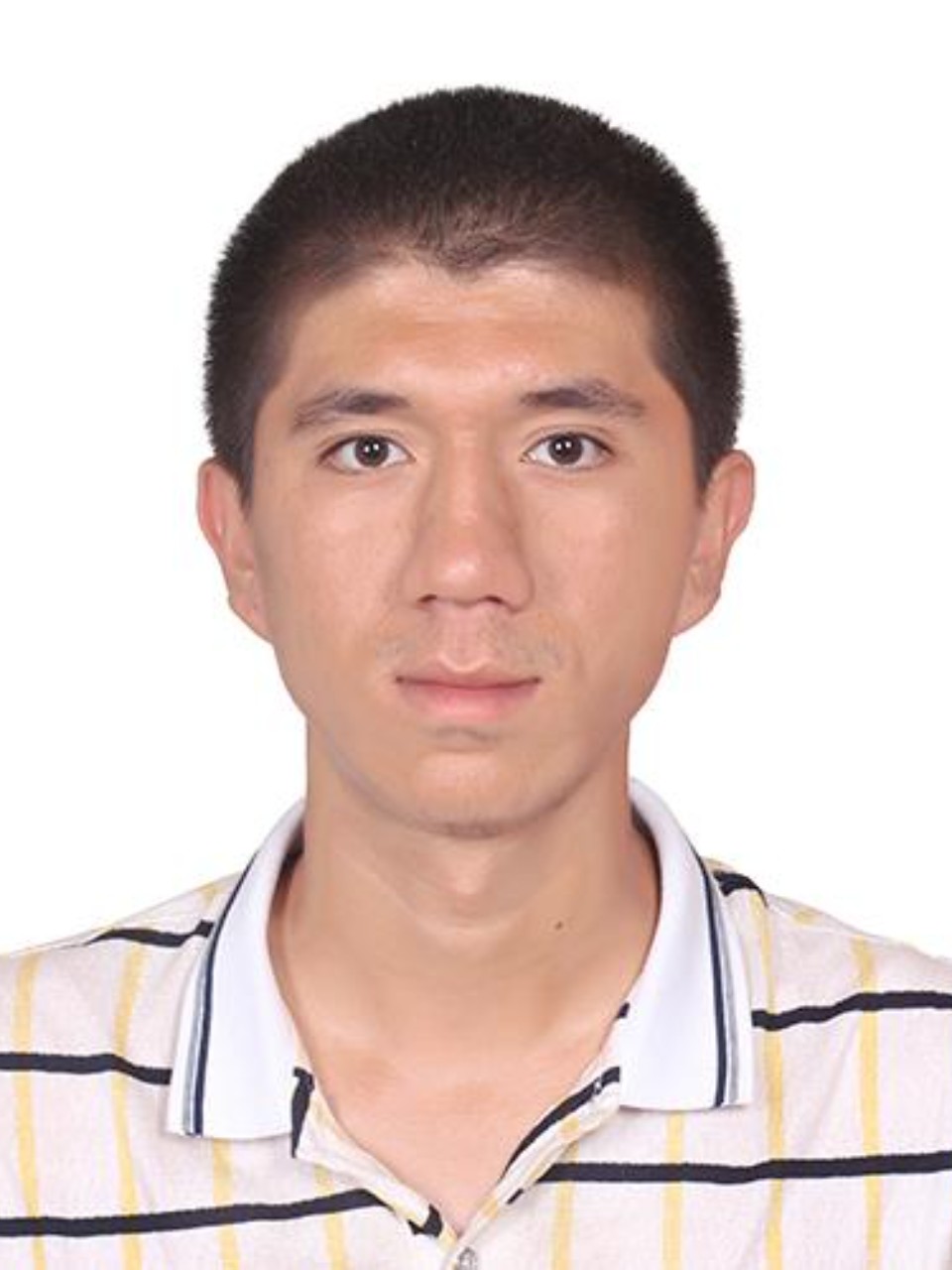}}]{Abudukelimu Wuerkaixi}
is currently pursuing the B.S. degree from the Department of Automation, Tsinghua University, Beijing, China. His research interests include machine learning and computer vision.
\end{IEEEbiography}

\begin{IEEEbiography}[{\includegraphics[width=1in,height=1.25in,clip,keepaspectratio]{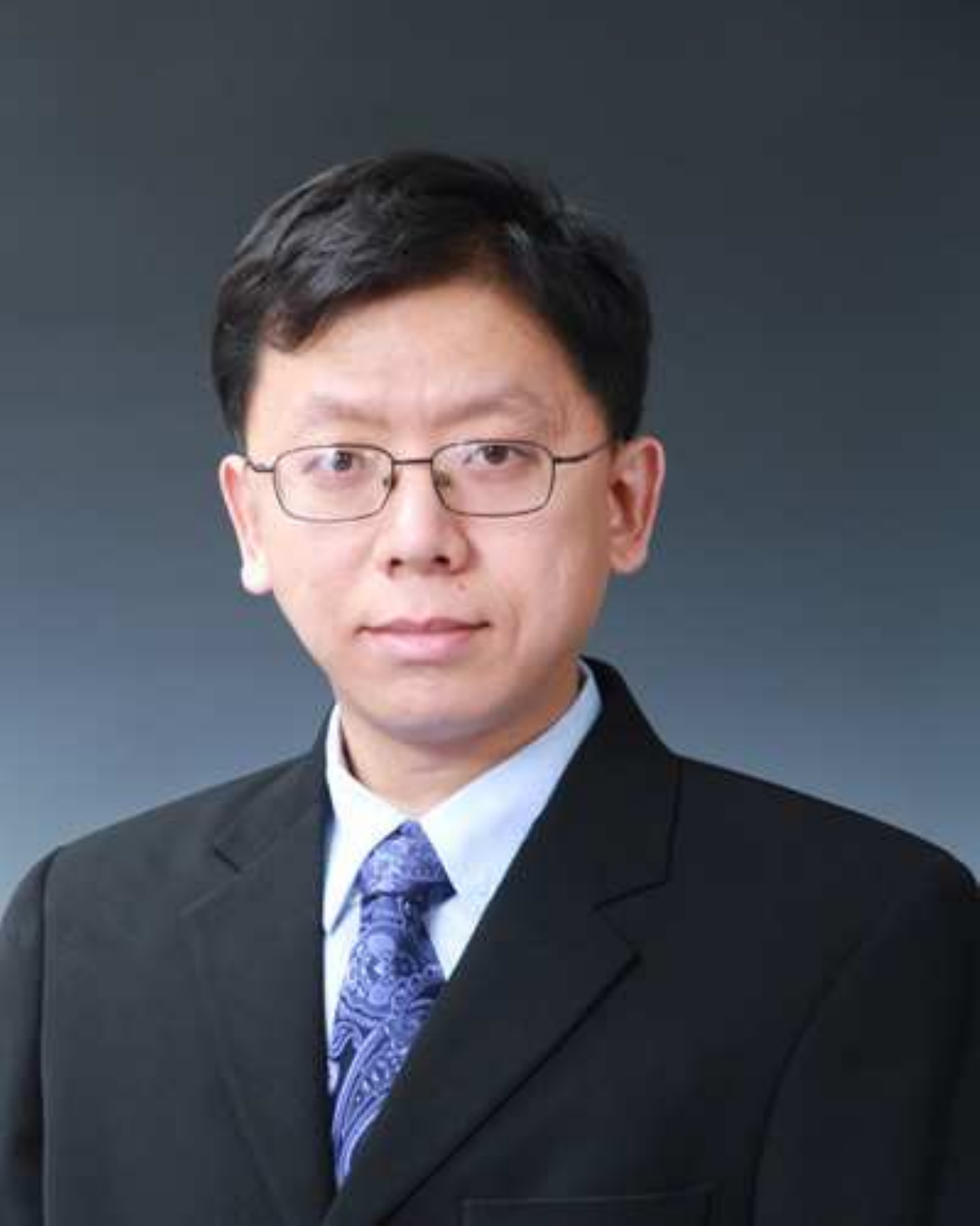}}]{Jianjiang Feng}
received the B.Eng. and Ph.D. degrees from the School of Telecommunication Engineering, Beijing University of Posts and Telecommunications, China, in 2000 and 2007, respectively. From 2008 to 2009, he was a Post-Doctoral Researcher with the PRIP Laboratory, Michigan State University. He is currently an Associate Professor with the Department of Automation, Tsinghua University, Beijing. His research interests include fingerprint recognition and computer vision. He is an Associate Editor of the Image and Vision Computing.
\end{IEEEbiography}

\begin{IEEEbiography}[{\includegraphics[width=1in,height=1.25in,clip,keepaspectratio]{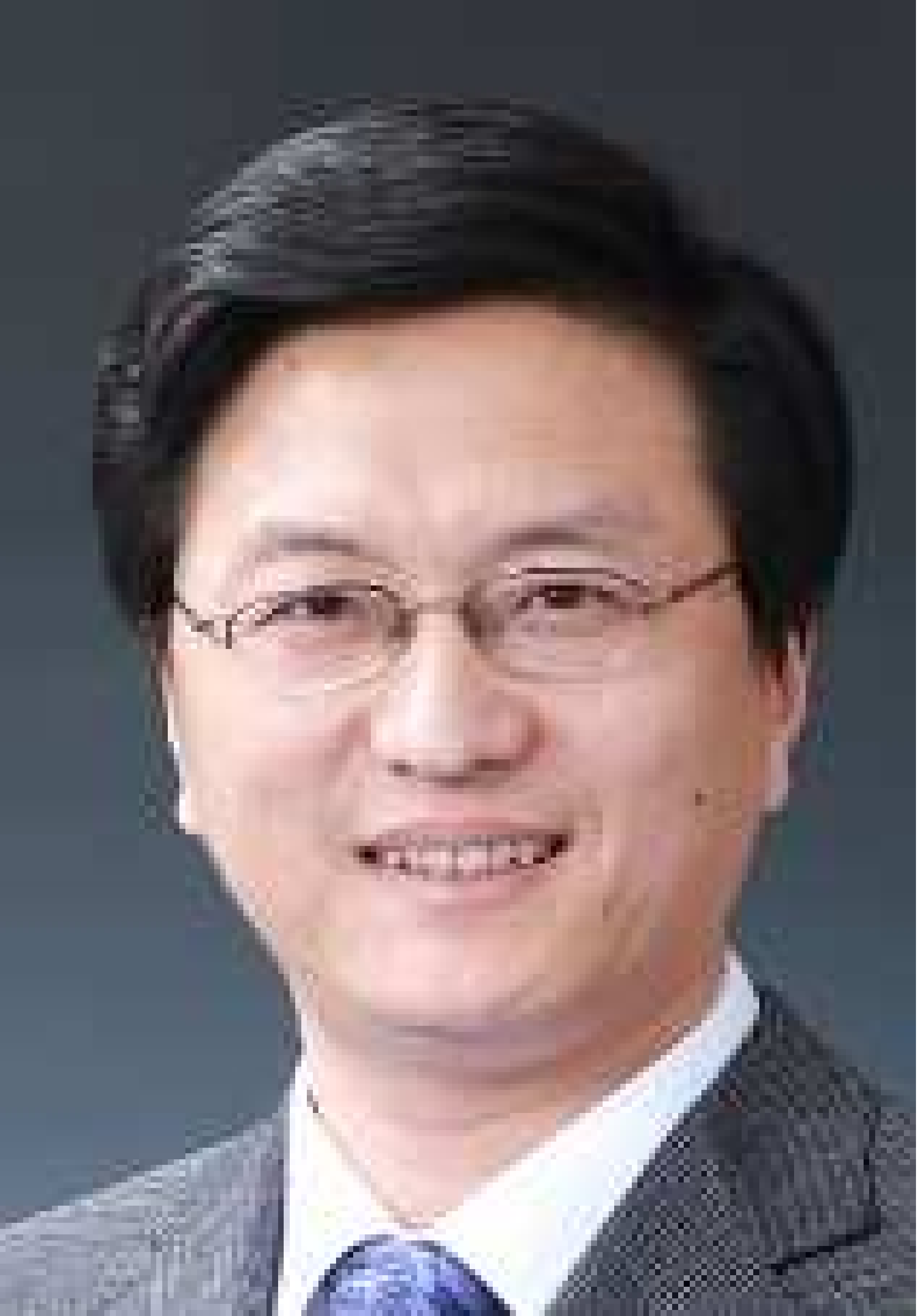}}]{Jie Zhou}
(M'01-SM'04) received the BS and MS degrees both from the Department of Mathematics, Nankai University, Tianjin, China, in 1990 and 1992, respectively, and the PhD degree from the Institute of Pattern Recognition and Artificial Intelligence, Huazhong University of Science and Technology (HUST), Wuhan, China, in 1995. From then to 1997, he served as a postdoctoral fellow in the Department of Automation, Tsinghua University, Beijing, China. Since 2003, he has been a full professor in the Department of Automation, Tsinghua University. His research interests include computer vision, pattern recognition, and image processing. In recent years, he has authored more than 100 papers in peer-reviewed journals and conferences. Among them, more than 30 papers have been published in top journals and conferences such as the IEEE Transactions on Pattern Analysis and Machine Intelligence, IEEE Transactions on Image Processing, and CVPR. He is an associate editor for the IEEE Transactions on Pattern Analysis and Machine Intelligence and two other journals. He received the National Outstanding Youth Foundation of China Award. He is an IAPR Fellow.\end{IEEEbiography}

\end{document}